%% file: main.tex
\PassOptionsToPackage{hyphens}{url}
\documentclass{article}

\usepackage[T1]{fontenc}  %
\usepackage{microtype}
\usepackage{graphicx}
\usepackage{wrapfig}  %
\usepackage{subcaption}
\usepackage{booktabs}
\usepackage{hyperref}
\usepackage{comment}
\usepackage{enumitem}

\usepackage[preprint]{icml2026}

\usepackage{amsmath}
\usepackage{amssymb}
\usepackage{mathtools}
\usepackage{amsthm}
\usepackage{xcolor}
\usepackage{colortbl}  %
\usepackage{tabularx}  %
\usepackage{siunitx}  %
\usepackage{nicematrix}  %
\usepackage{tikz}  %
\usepackage{rotating}  %
\usepackage{afterpage}  %
\usepackage[capitalize,noabbrev]{cleveref}

\input{macros}

\icmltitlerunning{Reducing Order Sensitivity in Decision Models}

\begin{document}

\twocolumn[
  \icmltitle{Reducing Order Sensitivity in Decision Models}
  \begin{icmlauthorlist}
    \icmlauthor{Bodong Liu}{cornell}
    \icmlauthor{Dongyoung Go}{cornell}
    \icmlauthor{Sofian Zalouk}{cornell} 
    \icmlauthor{Kilian Q. Weinberger}{cornell}
  \end{icmlauthorlist}
  \icmlaffiliation{cornell}{Department of Computer Science, Cornell University, New York, USA}
  \icmlcorrespondingauthor{Bodong Liu}{bl576@cornell.edu}
  \icmlkeywords{Machine Learning, order sensitivity, permutation invariance, positional bias, decision models}
  \vskip 0.3in
]

\printAffiliationsAndNotice{}

\input{00_abstract}
\input{01_introduction}

\input{03_definitions}
\input{02_motivation}

\input{04_methods}
\input{05_experiments}

\input{07_conclusion}
\input{99_ack}
\bibliography{references}
\bibliographystyle{icml2026}

\newpage
\appendix
\onecolumn
\input{R_related_work}
\input{A_metric_tables}
\input{B_method_details}
\input{C_datasets}
\input{D_setup}
\input{E_training}
\input{F_latency}
\input{G_hosted_api}
\input{H_precision}
\input{I_boxplots}
\input{J_separate_encoding}

\end{document}

%% file: macros.tex
\usepackage{xspace}
\usepackage{xfrac}
\newcommand{\para}[1]{\par\vspace{0.6em}\noindent\textbf{#1}}
\newcommand{\worstacc}{\ensuremath{\hat{A}_\mathrm{min}}}
\newcommand{\avgacc}{\ensuremath{\hat{A}_\mathrm{avg}}}
\newcommand{\theoryworstacc}{\ensuremath{A_\mathrm{min}}}
\newcommand{\theoryavgacc}{\ensuremath{A_\mathrm{avg}}}
\newcommand{\gap}{\ensuremath{\widehat{\Delta A}}}
\newcommand{\theorygap}{\ensuremath{\Delta A}}
\newcommand{\retentiontheory}{\ensuremath{{R}_A}}
\newcommand{\methodlong}{tree attention\xspace}
\newcommand{\method}{tree attention\xspace}
\newcommand{\Method}{Tree attention\xspace}

%% file: 00_abstract.tex
\begin{abstract}
Decision models are emerging as fast components of larger AI systems.
Yet recent decision models, such as Jev, process options sequentially, leading to different predictions based on their ordering. 
Across ten open- and closed-source decision models, we find that \emph{order-sensitivity} is pervasive, causing substantial accuracy degradation and unreliable decision-making. 
These effects become more pronounced as the number of options grows and can compound across multi-step decisions. 
To address this problem, we systematically compare eight methods for mitigating order sensitivity in terms of robustness, predictive accuracy, and inference efficiency. 
Among these methods, we find that \emph{tree attention} is a surprisingly effective and efficient method.
It achieves order invariance by processing options as parallel branches of a shared context.
When used during training, it matches the baseline's average accuracy.
Applied to the baseline without retraining, it also eliminates order sensitivity with only a 1.1-point drop in accuracy. 
Our findings demonstrate that order invariance can be enforced directly in decision models with little or no loss in predictive accuracy.

\end{abstract}

%% file: 01_introduction.tex
\section{Introduction}
\label{sec:intro}
\begin{figure*}[!t]
  \centering
  \includegraphics[width=\textwidth]{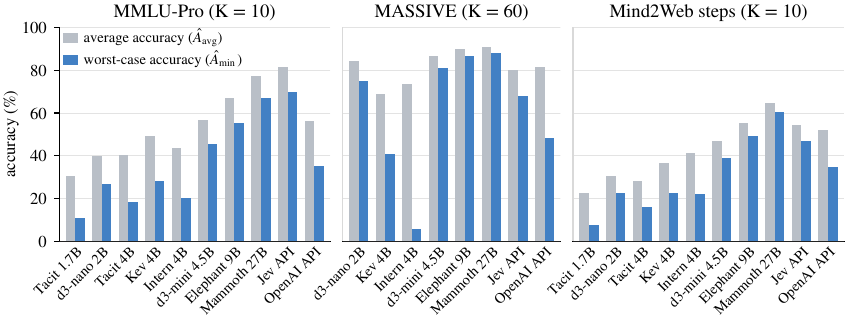}
  \caption{
  Existing models lose accuracy when the options are reordered.
  }
  \label{fig:1}
  \vspace{-10pt}
\end{figure*}

Recently, decision models have gained popularity for fast and typed decisions \citep{almeida2026jev,openai2026decisions}.
Given a context (question) and a set of candidate options, a decision model returns a distribution over the given options in a single prefill pass.
Therefore, they are well suited as lightweight components within larger systems, for example, routing documents, handling customer support, or choosing an agent's next tool call \citep{openai2026decisions,typesafe2026jackjill,ling2026jevwild,stefaniuk2026box,zhu2026jevagentharness}.
However, simply presenting the same options in a different order can change their decisions, even when the order is semantically irrelevant, a weakness known as \emph{order sensitivity}.

Order sensitivity has been widely studied in language models \citep{pezeshkpour2024order,zheng2023judging}, but it remains under-researched for decision models, where its consequences are arguably more severe.
In these systems, the downstream components use their decisions without human review, while the option order is set by upstream and can vary across calls.
The loss in performance grows with the option set size, which reaches thousands in industrial-scale applications \citep{zhang2025reic,qin2024toolllm}, and errors compound in multi-step decisions \citep{deng2023mind2web,sinha2026illusion}.
We investigate two questions: how pervasive and harmful is order sensitivity in existing models, and how we can remove it while preserving the key advantages of decision models.

First, we characterize the order sensitivity of ten open- and closed-source models by their worst-case performance (\cref{sec:motivation}).
We find that order sensitivity is both pervasive and harmful. On challenging knowledge questions, 
all models we evaluate
lose 13.5--64.2\% of their average accuracy in the worst case (\cref{fig:1}).
Furthermore, this effect grows with the number of options: on the same intent classification dataset, 
the OpenAI Decisions API's worst-case loss jumps from 2.2\% to 55.8\% as the number of options increases from 4 to 151.
Errors compound on multi-step decision tasks, with even the most robust model retaining only 53.8\% of its success rate by the fifth step (\cref{fig:2}).

Second, we systematically compare eight methods adapted from prior work that address order sensitivity.
We revisit methods from prior work spanning multiple categories: post-hoc calibration of the output distribution 
\citep{zhao2021calibrate,zheng2024selectors}%
, training-time augmentation with permuted orders \citep{pradeep2023rankvicuna,zhu2025judgelm}, test-time ensembling over orders \citep{wang2024fair}, and architectural designs \citep{mcilroy2024order,amit2026candidate}.
By implementing them on a shared baseline decision model, we judge each method by its average and worst-case accuracy, and its training and inference cost (\cref{sec:setup}).

We find that tree attention is a surprisingly efficient and effective method.
It offers order invariance in a single forward pass by processing options as parallel branches of a shared context.
It does not lower average accuracy when used in training and loses only 1.1 points when applied to the baseline at test time.
In comparison, post-hoc calibration and training-time augmentation have little effect; test-time ensembling costs multiple forward passes and scales poorly as the option set size increases; filtered ensembling suffers from high latency; and separate option encoding lowers average performance substantially.
Additionally, we trained a 4B model using \methodlong{} and show that its worst-case performance is better than or on par with existing models of comparable size (\cref{sec:sanity-check}).

%% file: 03_definitions.tex
\section{Definitions and Metrics}
\label{sec:defs}
A dataset has $N$ items (questions). Item $n$ consists of a context $x_n$ and a set of options $\mathcal{A}_n=\{a_1,\dots,a_K\}$ of size $K$, including a gold option $g_n\in\mathcal{A}_n$ (henceforth, we drop the index $n$ for simplicity).
An option order is a permutation $\pi$ of $\{1,\dots,K\}$ that lists option $a_{\pi(j)}$ at position $j$. 
Let $\Pi$ denote the set of all $K!$ possible option permutations.
A decision model $\theta$ reads $(x, a_{\pi(1)}, \dots, a_{\pi(K)})$ and returns a probability distribution $p_\theta(\cdot \mid x, \pi)$ over the options.
We extract the answer $\hat{a}(\pi)=\arg\max_{a\in\mathcal{A}} p_\theta(a\mid x, \pi)$.

\begin{figure*}[!t]
  \centering
  \includegraphics[width=\textwidth]{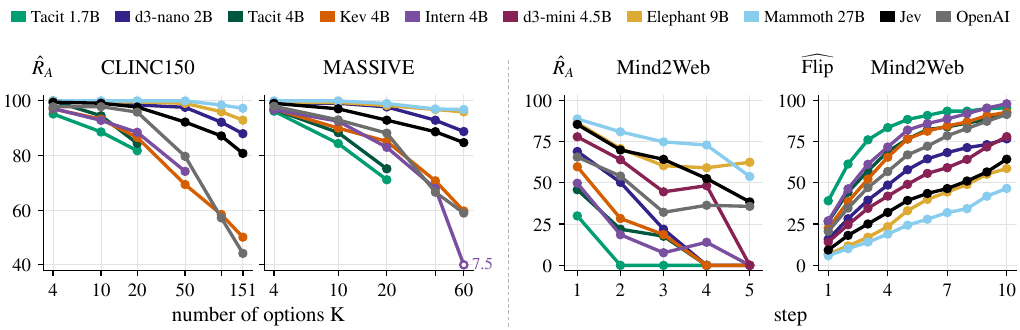}
  \caption{
  Order sensitivity grows with more options and over steps.
  Left two panels show that accuracy retention $\hat{R}_A$ ($\uparrow$) falls with the option count $K$. 
  On MASSIVE, Intern 4B at $K=60$ (7.5\%) lies far below the y-axis range and is drawn as a hollow marker.
  Right two panels show that order sensitivity compounds over multi-step decisions, indicated by decreasing $\hat{R}_A$ and growing flip rate ($\downarrow$) over steps.}
  
  \label{fig:2}
  \vspace{-10pt}
\end{figure*}
\label{sec:order-sensitivity}
\para{Order sensitivity} describes the extent to which a decision model's predictions or performance vary with the ordering of options when the ordering carries no semantic information.
Conversely, order invariance is the property that a model's prediction remains unchanged under permutations:
\begin{equation}
    p_\theta(a \mid x, \pi)=p_\theta(a\mid x, \pi') \quad \forall a \in \mathcal{A}, \pi, \pi'.
    \label{eq:invariance_distribution}
\end{equation}
We focus on the weaker but more practical property that the selected answer $\hat{a}(\pi)$ is invariant under permutation:
\begin{equation}
    \hat{a}(\pi)=\hat{a}(\pi') \quad \forall \pi, \pi'.
    \label{eq:invariance_answer}
\end{equation}

\vspace{-0.25em}
\para{Metrics.}
We first define all metrics over the full set of option orders $\Pi$, and then propose efficient estimators that we use in practice. These metrics capture the different effects of order sensitivity on robustness and performance.

\para{Average accuracy (\theoryavgacc)} measures the mean accuracy across 
option orders:
\begin{equation}
A_{\mathrm{avg}}=\frac{1}{N}\sum_{n=1}^{N} \frac{1}{|\Pi|}\sum_{\pi \in \Pi} \mathbf{1}\left[\hat{a}_{n}(\pi)=g_{n}\right].
\end{equation}

\para{Worst-case accuracy (\theoryworstacc)} measures the share of questions answered correctly under all 
option orders:
\begin{equation}
A_\mathrm{min}=\frac{1}{N}\sum_{n=1}^{N} 
\mathbf{1}\left[\hat{a}_{n}(\pi)=g_{n} \ \ \forall \pi \in 
\Pi\right].
\end{equation}

\para{Accuracy drop (\theorygap)} measures the degradation from average to worst-case accuracy:
\begin{equation}
\Delta A = A_{\mathrm{avg}}-A_{\min}.
\end{equation}

\para{Accuracy retention (\retentiontheory)} is the share of average accuracy kept under the worst-case order:
\begin{equation}
R_A = A_{\mathrm{min}} / A_{\mathrm{avg}}.
\end{equation}
We use this metric when average accuracy varies.

\para{Pairwise flip rate} measures the fraction of pairs of option orders that yield different answers:
\begin{equation}
\mathrm{Flip}=\frac{1}{N}\sum_{n=1}^{N} \binom{|\Pi|}{2}^{-1} \sum_{\{\pi, \pi'\} \subseteq \Pi} \mathbf{1}\left[\hat{a}_{n}(\pi)\neq\hat{a}_{n}
(\pi')\right].
\end{equation}

\para{Estimation.}
Evaluating all $K!$ orders is infeasible for large $K$.
For each item $n$, we therefore use a subsample $\Pi_n \subseteq \Pi$ consisting of the $K$ orders given by the rows of a random Latin square over the options, which counterbalances position and option: every option appears exactly once in every position.
We obtain the estimates $\hat{A}_{\mathrm{avg}}$, $\hat{A}_{\min}$, $\widehat{\Delta A}$, $\hat{R}_A$, and $\widehat{\mathrm{Flip}}$ by replacing $\Pi$ with $\Pi_n$ in the definitions above, i.e., by averaging over $\pi \in \Pi_n$ and over pairs $\pi_s, \pi_t \in \Pi_n$ with $s<t$.
All reported numbers are these estimates. 
Since $\Pi_n \subseteq \Pi$, our worst-case accuracy is an upper bound ($\hat{A}_{\min} \geq A_{\min}$), so our estimates understate the true order sensitivity (see \cref{tab:app-all-orders} for how tight the estimate is).

\begin{figure*}[!t]
  \vspace{-5pt}
  \centering
  \includegraphics[width=\textwidth]{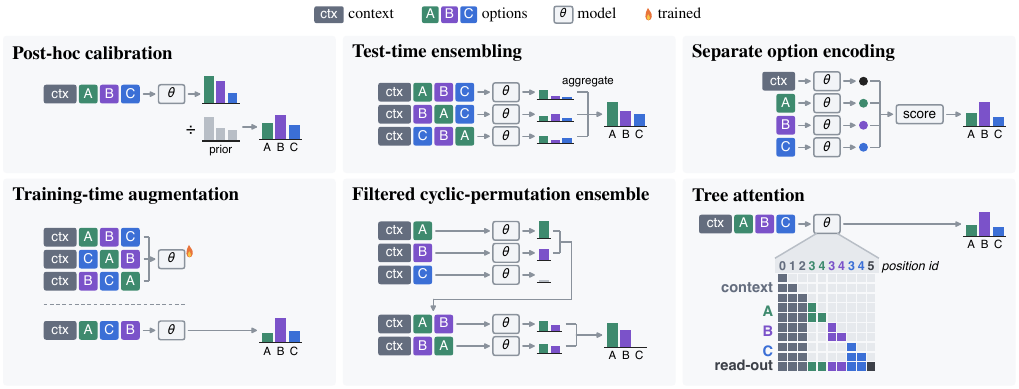}
  \caption{
Illustration of how different methods process one question.
  }
  \vspace{-10pt}
  \label{fig:3}
\end{figure*}

%% file: 02_motivation.tex
\section{Order Sensitivity in Existing Models}
\label{sec:motivation}

This section quantifies the order sensitivity of existing decision models.
We test 8 open-source decision models ranging from 1.7B to 27B parameters: Tacit 1.7B and 4B \citep{tacit2026_1.7b,tacit2026_4b}, d3-nano 2.2B \citep{vllmsr2026d3nano}, Kev 4B \citep{palmer2026kev4b}, Intern-Decision 4B \citep{zhu2026intern}, d3-mini 4.5B \citep{vllmsr2026d3mini}, and Elephant 9B and Mammoth 27B \citep{baselabs2026mammoth}. Of these, d3-nano, d3-mini, Elephant, and Mammoth are top-ranking in their size cohort on the Jev Decision Index \citep{decisionindex}. We also test two closed-source models: TypeSafe AI's Jev and OpenAI Decisions API.

Evaluation covers three task types representative of
decision model applications, using the metrics defined in \cref{sec:defs}: 
\begin{enumerate}[leftmargin=*, labelsep=4pt, itemindent=0pt, itemsep=2pt, parsep=0pt, topsep=0pt]
    \item Multiple-choice question answering: ARC \citep{clark2018arc}, MMLU \citep{hendrycks2021mmlu}, and MMLU-Pro \citep{wang2024mmlupro}.
    \item Intent classification: BANKING77 \citep{casanueva2020banking}, CLINC150 \citep{larson2019clinc}, and MASSIVE \citep{fitzgerald2023massive}.
    \item Multi-step web tasks: Mind2Web \citep{deng2023mind2web}.
\end{enumerate}

\para{Existing decision models suffer from order sensitivity.}
As shown in \cref{fig:1}, every tested model loses substantial accuracy under the worst-case order, with accuracy retention ranging from 35.8\% to 86.5\% on MMLU-Pro, 7.5\% to 96.8\% on MASSIVE, and 34.2\% to 93.2\% on Mind2Web.
Larger models tend to be more robust, but scaling the model size does not remove order sensitivity effectively: Mammoth 27B loses only about 3 points less than d3-nano 2.2B on MMLU-Pro and Mind2Web.

\para{The sensitivity grows with the option set size.}
To examine how sensitivity scales with the option set size, we vary the number of input options on CLINC150 and MASSIVE.
\cref{fig:2} shows that all systems start with little accuracy drop with 4 options, but many degrade sharply as early as when the option count rises to 10.
To highlight two severe cases: the OpenAI Decisions API keeps 97.8\% of its average accuracy with 4 options but only 44.2\% with all 151 options of CLINC150; Intern 4B keeps 96.8\% with 4 options but only 7.5\% with 60 options of MASSIVE.
This effect limits decision models deployed for settings with large option sets such as hundreds of intents, thousands of tools and APIs \citep{qin2024toolllm,patil2024gorilla}, or routing requests among $>$40,000 skills \citep{kim2018largescale}.

\para{The sensitivity compounds over steps.}
In multi-step tasks such as web navigation, a major application of decision models, small instabilities can surface and accumulate at every step; yet the effect of order sensitivity in such settings remains unexplored.
On Mind2Web, we test models on 190 tasks with at least five steps.
We find that the worst-case accuracy degrades over steps, and by the fifth step, eight of the ten models retain less than half of their average accuracy (\cref{fig:2}).
The most robust model in the previous tests, Mammoth 27B, retains 88.8\% of its average task success at step 1 but only 53.8\% by step 5.
We also measure the pairwise path flip rate, i.e., how often a model produces a different path when options are presented in a different order.
We find that the flip rate increases steadily: by step 10, nine of the ten models flip with more than a 50\% chance, and the most robust one flips its path with a 46.6\% chance.

Together, these observations suggest that evaluations of decision models should report their robustness by measuring performance across multiple option orders rather than a single fixed order.
The observations also call for methods that reduce order sensitivity.

%% file: 04_methods.tex
\section{Methods for Reducing Order Sensitivity}
\label{sec:methods}

In this section, we review representative methods for reducing order sensitivity.
We group them into four families according to where they intervene: post-hoc calibration, training-time augmentation, test-time ensembling, and architectural design.
We summarize these methods in \Cref{fig:3}, and provide a discussion of related work in Appendix~\ref{app:related}.

\subsection{Post-hoc calibration}
Calibration only changes how the $K$ output probabilities turn into an answer. 
Most methods rescale the log-probability of each position with parameters shared across items \citep{guo2017calibration,zhao2021calibrate}, estimated either without labels \citep{zhao2021calibrate,fei2023label,zhou2024batch,zheng2024selectors,lv2026emptiness} or from labelled data \citep{guo2017calibration}.
We test one method of each kind.

\para{Prior correction} (PriDe; \citealp{zheng2024selectors}) estimates the model's prior over positions from a few (typically 5\%) items answered in several orders and divides it out of every prediction.
While PriDe was originally proposed for the prior over option ID tokens, we apply it to positions.

\para{Per-position temperature scaling} assigns each position $k$ a temperature $\tau_k$ and takes the answer distribution as a softmax over $\log p_k/\tau_k$.
The temperatures are fitted on labeled items from the target distribution.

\subsection{Training-time augmentation}

A model can pick up position preferences during training, and many studies offset the bias by augmenting the training data.
Data augmentation shows each training example multiple times under different option orders \citep{pradeep2023rankvicuna,zhu2025judgelm}; Appendix~\ref{app:related} lists further applications.
We test one common form: training for several epochs where each epoch gives a new random option order for every question. 

\subsection{Test-time ensembling}

Ensembling asks the model to answer the same question under different option orders and aggregates the answers into one.
Typical aggregations include averaging the probabilities (\emph{avg}) \citep{zheng2024selectors,raghavan2026reflex}, majority vote (\emph{vote}) \citep{pezeshkpour2024order}, and taking a geometric mean \citep{anyjev2026}.
Averaging over all $K!$ orders makes the answer exactly order-invariant \citep{izacard2023atlas,zheng2024selectors}, but its factorial complexity makes it infeasible.
Methods differ mainly in which orders they use, and we test three.

\para{Random-permutation ensemble} aggregates the answers over $m$ random orders \citep{pezeshkpour2024order,hou2024rankers,tang2024found}.
It approaches the full $K!$ average as $m$ grows, but for finite $m$ it is order-invariant only in expectation over the draw. We report the random ensemble with avg and vote.

\para{Cyclic-permutation ensemble} presents the question in the $K$ cyclic shifts of a start order \citep{izacard2023atlas,zheng2024selectors}.
These shifts form a Latin square, the smallest set in which each option takes each position exactly once.
AnyJev \citep{anyjev2026} combines this design with the \emph{L0} reducer, which estimates the model's prior over positions from the evaluation set and divides it out of each answer before taking the geometric mean (Appendix~\ref{app:methods}).
We report the cyclic ensemble with both avg and L0.

\para{Filtered cyclic-permutation ensemble} first filters out a shortlist of options and then ensembles over the shortlist only
\citep{deng2023mind2web, tang2024found, zhu2025optionid, blankenstein2026biasbusters}.
We follow Runoff \citep{huang2026runoff} which formalizes this design for decision models.
At the filter stage, the model scores each option individually. 
The filter then selects a set $C$ of at least $k_{\min}$ top-$p$ options whose combined probability mass reaches $p$.
Since each option is scored independently, this stage is order-invariant.
At the ensemble stage, a cyclic-permutation ensemble over $C$ produces the final answer.
The method costs $|C|{+}1$ passes in two sequential steps, where $k_{\min}\le|C|\le K$.

\subsection{Architectural designs}
\para{Separate option encoding} encodes each option 
in isolation from the question and other options, and then symmetrically scores all option encodings  \citep{reimers2019sbert,karpukhin2020dpr,khattab2020colbert,humeau2020poly}.
Among decision models, Jeva \citep{li2026jeva} implements this by using a shared encoder to encode the request and options independently, and then scoring each option encoding against the question encoding symmetrically.
It guarantees order-invariance and allows individual option encodings to be cached and reused across requests.

\para{\Method} treats the options as parallel branches that share the context $x$ as a common prefix.
Given $x$ and options $a_1, \dots, a_K$, \Method modifies the attention mask so that each token in $a_k$ attends to $x$ and to the preceding tokens of $a_k$, but not to tokens of other options $a_{j \neq k}$ (\cref{fig:3}, bottom right).
Variants of this idea have been explored \citep{mcilroy2024order,brown2025finetuning,egressy2025set} and are most prominently applied to speculative decoding \citep{miao2024specinfer,cai2024medusa}.
Some recent work applies it to decision models \citep{von,biswas2026bevdecider,amit2026candidate}; we describe each in Appendix~\ref{app:related}.
Since each option is encoded independently of the others and at identical positions, the method is order-invariant by construction and requires a single forward pass for any $K$.

\enlargethispage{\baselineskip}
\subsection{Other approaches}
\label{sec:other-approaches}
We do not test other methods because they are less common or less practical for training a decision model.
Appendix~\ref{app:related} reviews the other related work in full.

%% file: 05_experiments.tex
\section{Experiments}
\input{table_systems.tex}
\label{sec:setup}
\vspace{-9pt}
\para{Baseline.}
To systematically compare the methods introduced in \cref{sec:methods}, we implement all of them on a shared decision model design that follows prior work \citep{devlin2019bert,kev}.
We fine-tune on Qwen3-1.7B-Base \citep{yang2025qwen3} and replace its language head with a read-out head that scores each option:
\begin{equation}
  p_\theta(a \mid x, \pi) = \operatorname{softmax}_a\big(\tau \cos(W_q h^{\pi}_{\mathrm{r}},\, W_k h^{\pi}_{a})\big),
  \label{eq:readout}
\end{equation}
where $h^{\pi}_{\mathrm{r}}$ and $h^{\pi}_{a}$ are the backbone's final hidden states at the read-out token and at the end of option $a$ under order $\pi$; $W_q$ and $W_k$ are learned projections; and $\tau$ is a learned temperature.
We call this design, and the model trained with it, \emph{baseline}; every other variant is either applied to baseline at test-time or trained under the same recipe.

The training corpus contains 891,107 examples from 48 public sources covering diverse domains (see training details in \cref{app:training}). 
All models are trained on the same copy preprocessed so that gold options are uniformly distributed at all positions.
This prevents models from picking up extra sensitivity from positional skew in the training set.

\para{Method variants.} Calibration and ensembling are applied post hoc to baseline as is.
Augmentation iterates the training data in full for every epoch where each gives a new option order for each question.
Since this introduces more update steps, we test a control trained for four epochs without permuting the options.
Separate encoding is trained using the baseline design.
As an ablation, we also train a variant using Jeva's proposed scoring head.
Tree attention modifies the backbone but not the baseline read-out head, and we test two variants: applying tree attention to the baseline's backbone at test time (without further training), and training a model with tree attention.

\para{Evaluation.}
We evaluate every arm on 7 datasets in 3 domains. See Appendix~\ref{app:datasets} for the datasets and Appendix~\ref{app:setup} for the full setup.
\begin{itemize}[nosep, leftmargin=*]
    \item \emph{Multiple choice:} ARC-Challenge ($N{=}1{,}165$),
MMLU ($N{=}1{,}909$),  and MMLU-Pro ($N{=}1{,}659$).
    \item \emph{Intent classification:}
BANKING77 ($N{=}300$),
CLINC150 ($N{=}200$),
and MASSIVE ($N{=}200$).
    \item \emph{Agentic:} Mind2Web ($N{=}2{,}118$).
\end{itemize}

\para{Cost.} We also measure the inference cost of each method because low-cost and fast response are key advantages of decision models. 
We report each method's token cost per request (relative to the baseline), latency, and throughput.
All methods are benchmarked under the same hardware and resource budget.
We stress test different request shapes: MMLU-Pro for short requests, CLINC150 for large option sets, and QuALITY \citep{pang2022quality} for long contexts (see details in Appendix~\ref{app:latency}).

\section{Results}
\subsection{Average and worst-case accuracy}
\label{sec:res-sensitivity}

\begin{figure*}[!t]
  \centering
  \includegraphics[width=\textwidth]{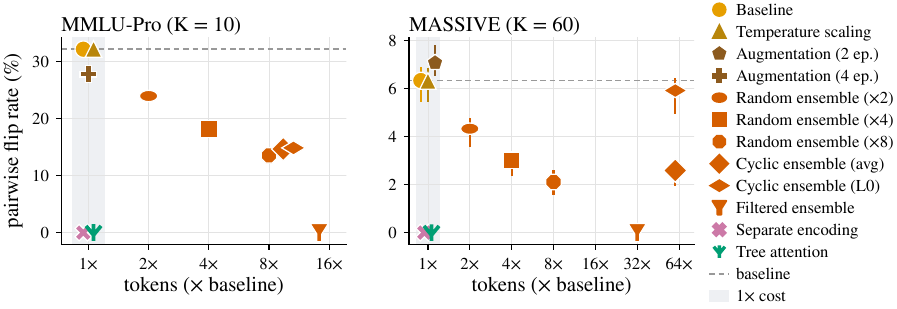}
  \caption{
  Pairwise flip rate ($\downarrow$) vs. inference token cost ($\downarrow$) relative to baseline. 
  Marker = mean over 3 training seeds; bar = min--max over the seeds; dashed line = baseline. For calibration and augmentation, only the best method of each family is shown in each panel.
  Markers in the shadowed area all have $1\times$ the token cost.
  }
  \vspace{-10pt}
  \label{fig:4}
\end{figure*}

\cref{tab:1} compares the performance of method variants on different evaluation datasets (Appendix~\ref{app:tables} reports every metric for every system).
\cref{fig:4} shows the token cost and flip rate tradeoff.
Except for augmentation, separate encoding, and \method{} applied at test time, all methods' average accuracies are not significantly lower than the baseline's.

\para{Post-hoc calibration} barely reduces order sensitivity.
Averaged over the 7 datasets (\cref{tab:1}, last column), baseline experiences a 10.4 points accuracy drop; temperature scaling does not improve it, and PriDe raises the drop to 12.5.
PriDe widens the accuracy gap mostly on the three small datasets (MASSIVE, BANKING77, and CLINC150) likely because a prior estimated from smaller sets is noisy and harmful.
This suggests that the baseline's order sensitivity comes mainly from question-specific effects rather than a global position bias.
Both calibration methods learn one correction per position shared by all questions, so they can only remove a global position preference and leave the question-specific sensitivity unchanged.

\para{Training-time augmentation} reduces order sensitivity slightly but lowers OOD average accuracy.
It lowers the accuracy drop by under one point from the baseline (from 10.4 points to 9.8 with 2 ep. and 9.5 with 4 ep.).
We find that the augmentation variants lower the average accuracy on the OOD sets with a clear trend: baseline (59.5)$>$ 2 ep. (58.7) $>$ 4 ep. (57.1).
The fixed-order control, trained for four epochs but in the same option order, has the largest accuracy drop and the lowest OOD average accuracy among the augmentation arms.
The pattern is consistent with overfitting on repeated data.
Together, these results suggest that exposure to new orders helps reduce sensitivity, but the additional epochs also raise the risk of overfitting, which may absorb most of the reduction and result in a modest net reduction.

\para{Test-time ensembling} reduces the accuracy drop in proportion to the extra number of passes.
With 2, 4, and 8 passes, the random ensemble reduces the accuracy drop from 10.4 points down to 7.0, 5.0, and 3.5 points, respectively.
On MMLU-Pro with $K{=}10$, ensembling 8 or 10 passes still leaves the worst-case accuracy 7.8--8.8 points below the average accuracy.
On sets with more than 8 options, the cyclic ensemble covers many more orders than the random ensemble ($\times 8$) but instead suffers a larger accuracy drop (e.g.\ 8.7 vs.\ 7.8 points on MMLU-Pro).
This is likely because cyclic ensemble rotates the options instead of completely re-ordering, and that may keep some bias from the fixed relative order.
Among the ensembling family, only the filtered ensemble is order-invariant.
As expected for ensembling, all of them raise average accuracy slightly (up to 1.7 points, on MMLU-Pro).

\para{Separate encoding} is order-invariant but has the lowest average accuracy of all compared methods.
It performs 7.4 points lower than the baseline on all sets and 10.7 points lower on the OOD sets, and the variant trained with Jeva's proposed scoring head falls 9.6 and 14.2 points below the baseline.
Its poor performance is not a surprise. 
Because the option tokens cannot attend to context tokens, the model may miss the fine details that dictate the correct answer.
For example, we find this method struggles on negated questions (``which is NOT ...''), the subset on which it achieves substantially lower accuracy than the baseline (44.4\% vs. 68.0\%).
We cover more details on discovered failure modes in Appendix~\ref{app:sep}.
\para{Tree attention} is fully order-invariant.
By construction, the tree attention mask keeps the options isolated from one another, and forcing each option to start from the same position index prevents the model from inferring a position for an option.
Each option's hidden state is therefore identical under every order, as shown by both variants attaining an accuracy drop of 0.0 on every dataset.
The trained variant performs the same as the baseline across all datasets.
On the other hand, applying tree attention to the baseline without further training lowers average accuracy by 0.9 points on the OOD sets and 1.1 points on all sets.
This slight decrement is expected because the model is trained on plain attention only.
Although tree attention does not allow attention between options, it does not have the same catastrophic effect on average accuracy as separate encoding because the option tokens can attend to context tokens, and the read-out token attends to all tokens.

\subsection{Cost}
\label{sec:res-cost}
\para{Training cost.}
Calibration and ensembling are applied to baseline without further training.
Augmentation multiplies training compute by its number of epochs.
Separate encoding takes about $3.14\times$ the time per training step as the baseline, because each example is a padded batch of $K{+}1$ sequences that makes training less efficient.
Tree attention (trained version) takes about $1.20\times$ the time per step because its attention mask is executed as several kernel calls per layer (measurement details in Appendix~\ref{app:training}).

\para{Inference cost.}
\Cref{fig:4} shows each method's tradeoff between token cost and flip rate. 
For non order-invariant methods, they roughly follow a negative log-linear trend: doubling the tokens for a constant improvement.
Among the three order-invariant methods, tree attention and separate encoding cost the same amount of tokens as the baseline, while the filtered ensemble costs 14--32 times as many.

\input{table5.tex}
\cref{tab:latency} shows the latency and throughput comparison.
Ensembles pay the inference cost proportional to the additional passes.
Their latency increases as the extra passes saturate the accelerator's arithmetic units.
Their throughput falls roughly in proportion to the number of passes.
Cyclic ensemble runs the slowest on CLINC150 ($K{=}151$) because it needs to ensemble as many passes as the option count.
Although filtered ensemble's shortlist makes it cheaper than the cyclic ensemble at a large $K$, it needs two sequential steps that make it the slowest on other workloads.
Overall, ensembling multiplies the computation per question, which offsets the efficiency that makes decision models attractive.
Separate encoding does not cost more tokens but still needs to batch $K{+}1$ sequences with padding for each request, which draws down its efficiency. 
\Method{} has the lowest latency and the highest throughput of all methods on every request shape because it only needs one forward pass, suggesting that it is the most suitable method for both responsiveness and high-load serving.

\subsection{Scaling to large option sets}
\label{sec:exp-k}
\cref{tab:massive-k} shows how non-order-invariant methods scale with increasing option set size.
The baseline's accuracy drop grows from 0.8 points at $K{=}4$ to 11.0 points at $K{=}60$.
Augmentation in fact performs worse at each option count by raising the accuracy drop (except for the 2 ep. variant when $K{=}4$).
Ensembles shrink the drop at every $K$ but do not stop it from growing.
For example, random ensemble ($\times8$) and cyclic ensemble have an accuracy drop of 0.1 points when $K{=}4$, but the drop grow to 2.8 and 3.9 points when $K{=}60$, respectively.
The increasing trend indicates that ensembling, while requiring multiples of inference cost, cannot reliably handle an increasing option set.

\subsection{Comparison with open models}
\input{table_massive_k.tex}
\label{sec:sanity-check}
To examine whether our findings extend to released models and to larger scale, we compare our 1.7B baseline and \method{} models, together with a 4B \method{} model, against released decision models of comparable size (\cref{tab:released}).
As these models differ from the ones we train in data and recipe, we treat this comparison as complementary to the controlled experiments above.

\suppressfloats[t]
\input{table_released.tex}

\Cref{tab:released} shows that all but Jeva lose accuracy under the worst-case order.
\Method{}, in contrast, attains the highest worst-case accuracy in each size group, or one statistically indistinguishable from it.
At 1.7B, it trails the slightly larger d3-nano (2.2B) by 2.3 points in average accuracy over all sets, but exceeds it by 6.2 points in worst-case accuracy (67.9 vs.\ 61.7).
At 4B, it outperforms Kev, Jeva, and Intern-Decision in both average and worst-case accuracy, including a 10.2-point gain in average accuracy over all sets over Jeva, the only other order-invariant model.
The model d3-mini achieves higher average accuracy, but its advantage vanishes under the worst-case order, where the two models are not significantly different.
Finally, scaling \method{} from 1.7B to 4B improves accuracy by 6.4 points over all sets while invariance is preserved, suggesting that the method also works at a larger scale.
Overall, these results indicate that \method{} remains competitive with released models of comparable size while guaranteeing order invariance.

%% file: table_systems.tex
\begin{table*}[!t]
\definecolor{invrow}{RGB}{222,235,250}\definecolor{lowcell}{RGB}{252,222,222}
\centering
\caption{Average accuracy (\avgacc{} $\uparrow$) and accuracy drop (\gap{} $\downarrow$), in \%. 
Mean (OOD): average over the 3 out-of-domain sets; Mean (all): over all 7; n/a if a set is missing. 
{\setlength{\fboxsep}{0.8pt}\colorbox{invrow}{Blue}}: order-invariant. {\setlength{\fboxsep}{0.8pt}\colorbox{lowcell}{Red}}: significantly worse than the baseline model (95\% paired interval). 
Mean over $n{=}3$ training seeds. 
$^\star$ out-of-domain.
n/a: not run, since evaluating the cyclic ensemble takes $K^2$ passes per question (Appendix~\ref{app:setup}).
}
\label{tab:1}\label{tab:systems}
\scriptsize
\setlength{\tabcolsep}{3pt}
\NiceMatrixOptions{custom-line={command=nmdashrule, tikz={dash pattern=on 1.5pt off 1.5pt}, total-width=0.3pt}}
\begin{NiceTabular}{@{}l@{\hspace{9pt}}w{c}{13pt}w{c}{13pt}Xw{c}{13pt}w{c}{13pt}Xw{c}{13pt}w{c}{13pt}Xw{c}{13pt}w{c}{13pt}Xw{c}{13pt}w{c}{13pt}Xw{c}{13pt}w{c}{13pt}Xw{c}{13pt}w{c}{13pt}Xw{c}{13pt}w{c}{13pt}Xw{c}{13pt}w{c}{13pt}@{}}[width=\textwidth, create-extra-nodes]
\toprule
 & \multicolumn{2}{c}{\makebox[0pt]{MMLU$^\star$}} & & \multicolumn{2}{c}{\makebox[0pt]{MMLU-Pro$^\star$}} & & \multicolumn{2}{c}{\makebox[0pt]{MASSIVE$^\star$}} & & \multicolumn{2}{c}{\makebox[0pt]{ARC}} & & \multicolumn{2}{c}{\makebox[0pt]{BANKING77}} & & \multicolumn{2}{c}{\makebox[0pt]{CLINC150}} & & \multicolumn{2}{c}{\makebox[0pt]{Mind2Web}} & & \multicolumn{2}{c}{\makebox[0pt]{Mean (OOD)}} & & \multicolumn{2}{c}{\makebox[0pt]{Mean (all)}} \\
\cmidrule(lr){2-3}\cmidrule(lr){5-6}\cmidrule(lr){8-9}\cmidrule(lr){11-12}\cmidrule(lr){14-15}\cmidrule(lr){17-18}\cmidrule(lr){20-21}\cmidrule(lr){23-24}\cmidrule(lr){26-27}
System & {\makebox[0pt]{\scriptsize\avgacc}} & {\makebox[0pt]{\scriptsize\gap}} & & {\makebox[0pt]{\scriptsize\avgacc}} & {\makebox[0pt]{\scriptsize\gap}} & & {\makebox[0pt]{\scriptsize\avgacc}} & {\makebox[0pt]{\scriptsize\gap}} & & {\makebox[0pt]{\scriptsize\avgacc}} & {\makebox[0pt]{\scriptsize\gap}} & & {\makebox[0pt]{\scriptsize\avgacc}} & {\makebox[0pt]{\scriptsize\gap}} & & {\makebox[0pt]{\scriptsize\avgacc}} & {\makebox[0pt]{\scriptsize\gap}} & & {\makebox[0pt]{\scriptsize\avgacc}} & {\makebox[0pt]{\scriptsize\gap}} & & {\makebox[0pt]{\scriptsize\avgacc}} & {\makebox[0pt]{\scriptsize\gap}} & & {\makebox[0pt]{\scriptsize\avgacc}} & {\makebox[0pt]{\scriptsize\gap}} \\
\midrule
Baseline model & 62.4 & 9.1 & & 38.7 & 17.1 & & 77.5 & 11.0 & & 85.0 & 4.9 & & 88.9 & 7.9 & & 89.6 & 10.7 & & 33.6 & 12.3 & & 59.5 & 12.4 & & 67.9 & 10.4 \\
\addlinespace[2pt]\nmdashrule\addlinespace[2pt]
\multicolumn{27}{l}{\textit{Post-hoc calibration}} \\
PriDe (5\%) & 62.6 & 9.3 & & 38.7 & 17.8 & & 77.5 & 15.2 & & 85.0 & 5.0 & & 88.8 & 11.4 & & 89.2 & 16.9 & & 33.6 & 11.9 & & 59.6 & 14.1 & & 67.9 & 12.5 \\
Per-position temp. & 62.5 & 9.1 & & 38.8 & 17.3 & & 77.5 & 11.2 & & 85.0 & 5.0 & & 88.9 & 7.9 & & 89.6 & 10.7 & & 33.5 & 12.0 & & 59.6 & 12.5 & & 68.0 & 10.4 \\
\addlinespace[2pt]\nmdashrule\addlinespace[2pt]
\multicolumn{27}{l}{\textit{Training-time augmentation}} \\
Augmentation (2 ep.) & 61.5 & 8.2 & & 37.8 & 15.5 & & 76.9 & 11.8 & & 84.9 & 4.9 & & 91.5 & 6.7 & & 92.0 & 10.8 & & 35.6 & 10.9 & & 58.7 & 11.8 & & 68.6 & 9.8 \\
Augmentation (4 ep.) & 61.0 & 7.6 & & 35.1 & 13.3 & & 75.3 & 15.1 & & 84.1 & 3.9 & & 91.8 & 7.3 & & 91.8 & 9.6 & & 34.5 & 9.6 & & 57.1 & 12.0 & & 67.7 & 9.5 \\
Augmentation (4 ep.; fixed) & 60.6 & 11.9 & & 36.0 & 18.9 & & 74.0 & 19.5 & & 83.9 & 7.1 & & 90.9 & 11.9 & & 91.3 & 16.1 & & 33.1 & 13.7 & & 56.9 & 16.8 & & 67.1 & 14.2 \\
\addlinespace[2pt]\nmdashrule\addlinespace[2pt]
\multicolumn{27}{l}{\textit{Test-time ensembling}} \\
Random ensemble ($\times 2$) & 62.8 & 5.9 & & 39.5 & 13.3 & & 77.8 & 6.8 & & 85.4 & 3.2 & & 89.0 & 4.5 & & 89.8 & 6.7 & & 33.9 & 8.5 & & 60.0 & 8.6 & & 68.3 & 7.0 \\
Random ensemble ($\times 4$) & 63.0 & 4.3 & & 40.0 & 10.3 & & 77.9 & 4.1 & & 85.6 & 2.2 & & 89.1 & 3.3 & & 90.0 & 4.7 & & 34.1 & 6.2 & & 60.3 & 6.2 & & 68.5 & 5.0 \\
Random ensemble ($\times 8$) & 63.2 & 2.7 & & 40.4 & 7.8 & & 77.9 & 2.8 & & 85.8 & 1.3 & & 89.1 & 2.3 & & 90.1 & 3.3 & & 34.2 & 4.4 & & 60.5 & 4.4 & & 68.7 & 3.5 \\
Cyclic ensemble (avg) & 63.2 & 2.5 & & 40.4 & 8.7 & & 78.0 & 3.9 & & 85.6 & 1.4 & & {\textit{n/a}} & {\textit{n/a}} & & {\textit{n/a}} & {\textit{n/a}} & & 34.1 & 5.8 & & 60.5 & 5.0 & & {\textit{n/a}} & {\textit{n/a}} \\
Cyclic ensemble (L0) & 63.1 & 2.6 & & 40.2 & 8.8 & & 77.7 & 11.7 & & 85.6 & 1.8 & & {\textit{n/a}} & {\textit{n/a}} & & {\textit{n/a}} & {\textit{n/a}} & & 34.1 & 5.8 & & 60.4 & 7.7 & & {\textit{n/a}} & {\textit{n/a}} \\
Filtered ensemble & 63.2 & 0.0 & & 39.8 & 0.0 & & 77.8 & 0.0 & & 85.4 & 0.0 & & 89.4 & 0.0 & & 89.8 & 0.0 & & 34.0 & 0.0 & & 60.3 & 0.0 & & 68.5 & 0.0 \\
\addlinespace[2pt]\nmdashrule\addlinespace[2pt]
\multicolumn{27}{l}{\textit{Architectural designs}} \\
Separate encoding & 51.4 & 0.0 & & 28.8 & 0.0 & & 66.3 & 0.0 & & 71.6 & 0.0 & & 90.6 & 0.0 & & 85.5 & 0.0 & & 29.4 & 0.0 & & 48.8 & 0.0 & & 60.5 & 0.0 \\
Separate enc. (Jeva head) & 48.9 & 0.0 & & 27.4 & 0.0 & & 59.7 & 0.0 & & 68.7 & 0.0 & & 90.9 & 0.0 & & 87.0 & 0.0 & & 25.2 & 0.0 & & 45.3 & 0.0 & & 58.3 & 0.0 \\
\Method{} (test-time) & 61.4 & 0.0 & & 37.4 & 0.0 & & 77.2 & 0.0 & & 84.6 & 0.0 & & 86.9 & 0.0 & & 87.3 & 0.0 & & 32.7 & 0.0 & & 58.6 & 0.0 & & 66.8 & 0.0 \\
\Method{} & 62.4 & 0.0 & & 37.9 & 0.0 & & 76.7 & 0.0 & & 85.2 & 0.0 & & 89.6 & 0.0 & & 90.3 & 0.0 & & 33.0 & 0.0 & & 59.0 & 0.0 & & 67.9 & 0.0 \\
\bottomrule
\CodeAfter
\providecommand{\cellbox}[2]{\tikz\fill[#1, blend mode=multiply] ([xshift=-3pt]#2-medium.north west) rectangle ([xshift=3pt]#2-medium.south east);}
\cellbox{lowcell}{5-6}
\cellbox{lowcell}{5-9}
\cellbox{lowcell}{5-15}
\cellbox{lowcell}{5-18}
\cellbox{lowcell}{5-24}
\cellbox{lowcell}{5-27}
\cellbox{lowcell}{6-20}
\cellbox{lowcell}{8-2}
\cellbox{lowcell}{8-5}
\cellbox{lowcell}{8-23}
\cellbox{lowcell}{9-2}
\cellbox{lowcell}{9-5}
\cellbox{lowcell}{9-8}
\cellbox{lowcell}{9-9}
\cellbox{lowcell}{9-23}
\cellbox{lowcell}{10-2}
\cellbox{lowcell}{10-3}
\cellbox{lowcell}{10-5}
\cellbox{lowcell}{10-6}
\cellbox{lowcell}{10-8}
\cellbox{lowcell}{10-9}
\cellbox{lowcell}{10-11}
\cellbox{lowcell}{10-12}
\cellbox{lowcell}{10-15}
\cellbox{lowcell}{10-18}
\cellbox{lowcell}{10-21}
\cellbox{lowcell}{10-23}
\cellbox{lowcell}{10-24}
\cellbox{lowcell}{10-26}
\cellbox{lowcell}{10-27}
\cellbox{lowcell}{19-2}
\cellbox{lowcell}{19-5}
\cellbox{lowcell}{19-8}
\cellbox{lowcell}{19-11}
\cellbox{lowcell}{19-17}
\cellbox{lowcell}{19-20}
\cellbox{lowcell}{19-23}
\cellbox{lowcell}{19-26}
\cellbox{lowcell}{20-2}
\cellbox{lowcell}{20-5}
\cellbox{lowcell}{20-8}
\cellbox{lowcell}{20-11}
\cellbox{lowcell}{20-20}
\cellbox{lowcell}{20-23}
\cellbox{lowcell}{20-26}
\cellbox{lowcell}{21-2}
\cellbox{lowcell}{21-5}
\cellbox{lowcell}{21-14}
\cellbox{lowcell}{21-17}
\cellbox{lowcell}{21-20}
\cellbox{lowcell}{21-23}
\cellbox{lowcell}{21-26}
\cellbox{invrow}{17-3}
\cellbox{invrow}{17-6}
\cellbox{invrow}{17-9}
\cellbox{invrow}{17-12}
\cellbox{invrow}{17-15}
\cellbox{invrow}{17-18}
\cellbox{invrow}{17-21}
\cellbox{invrow}{17-24}
\cellbox{invrow}{17-27}
\cellbox{invrow}{19-3}
\cellbox{invrow}{19-6}
\cellbox{invrow}{19-9}
\cellbox{invrow}{19-12}
\cellbox{invrow}{19-15}
\cellbox{invrow}{19-18}
\cellbox{invrow}{19-21}
\cellbox{invrow}{19-24}
\cellbox{invrow}{19-27}
\cellbox{invrow}{20-3}
\cellbox{invrow}{20-6}
\cellbox{invrow}{20-9}
\cellbox{invrow}{20-12}
\cellbox{invrow}{20-15}
\cellbox{invrow}{20-18}
\cellbox{invrow}{20-21}
\cellbox{invrow}{20-24}
\cellbox{invrow}{20-27}
\cellbox{invrow}{21-3}
\cellbox{invrow}{21-6}
\cellbox{invrow}{21-9}
\cellbox{invrow}{21-12}
\cellbox{invrow}{21-15}
\cellbox{invrow}{21-18}
\cellbox{invrow}{21-21}
\cellbox{invrow}{21-24}
\cellbox{invrow}{21-27}
\cellbox{invrow}{22-3}
\cellbox{invrow}{22-6}
\cellbox{invrow}{22-9}
\cellbox{invrow}{22-12}
\cellbox{invrow}{22-15}
\cellbox{invrow}{22-18}
\cellbox{invrow}{22-21}
\cellbox{invrow}{22-24}
\cellbox{invrow}{22-27}
\end{NiceTabular}
\vspace{-10pt}
\end{table*}

%% file: table5.tex
\begin{table}[!b]
\vspace{-3pt}
\centering
\caption{Latency and throughput on one RTX A6000 (memory-efficient SDPA, bf16). p95, p50: latency in ms ($\downarrow$). q/s: throughput in questions per second ($\uparrow$). \textbf{Bold}: the best except the baseline. 
}
\label{tab:latency}
\scriptsize
\setlength{\tabcolsep}{2.2pt}
\NiceMatrixOptions{custom-line={command=nmdashrule, tikz={dash pattern=on 1.5pt off 1.5pt}, total-width=0.3pt}}
\begin{NiceTabular*}{\columnwidth}{@{\extracolsep{\fill}}lccccccccc@{}}
\toprule
 & \multicolumn{3}{c}{MMLU-Pro} & \multicolumn{3}{c}{CLINC150} & \multicolumn{3}{c}{QuALITY} \\
\cmidrule(lr){2-4}\cmidrule(lr){5-7}\cmidrule(lr){8-10}
Method & p95 & p50 & q/s & p95 & p50 & q/s & p95 & p50 & q/s \\
\midrule
Baseline model & 37.0 & 35.5 & 41.8 & 46.4 & 41.0 & 29.7 & 502 & 413 & 1.71 \\
\addlinespace[2pt]\nmdashrule\addlinespace[2pt]
Random ensemble (2) & 49.7 & 46.9 & 23.7 & 82.1 & 81.1 & 14.6 & 974 & 801 & 1.37 \\
Random ensemble (4) & 91.1 & 48.5 & 14.0 & 136 & 135 & 7.38 & 1969 & 1617 & 0.69 \\
Random ensemble (8) & 159 & 81.3 & 8.54 & 273 & 268 & 3.63 & 4042 & 3335 & 0.34 \\
Cyclic ensemble & 189 & 90.2 & 7.51 & 4945 & 4785 & 0.20 & 2074 & 1691 & 0.68 \\
Filtered ensemble & 299 & 138 & 4.84 & 4197 & 2768 & 0.36 & 4851 & 4172 & 0.28 \\
\addlinespace[2pt]\nmdashrule\addlinespace[2pt]
Separate encoding & 84.9 & 39.2 & 14.4 & 215 & 170 & 5.38 & 3554 & 2783 & 0.44 \\
Separate enc. (Jeva head) & 85.9 & 40.4 & 14.4 & 217 & 172 & 5.35 & 3562 & 2785 & 0.44 \\
\Method{} & \textbf{43.8} & \textbf{38.7} & \textbf{44.3} & \textbf{45.1} & \textbf{44.3} & \textbf{29.4} & \textbf{773} & \textbf{598} & \textbf{1.71} \\
\bottomrule
\end{NiceTabular*}
\vspace{-10pt}
\end{table}

%% file: table_massive_k.tex
\begin{table}[t]
\vspace{-3pt}
\centering
\caption{Accuracy drop (\gap{} $\downarrow$) on MASSIVE by number of options, in points. 
Mean over $n{=}3$ training seeds.
}
\label{tab:massive-k}
\scriptsize
\setlength{\tabcolsep}{3pt}
\NiceMatrixOptions{custom-line={command=nmdashrule, tikz={dash pattern=on 1.5pt off 1.5pt}, total-width=0.3pt}}
\begin{NiceTabular*}{\columnwidth}{@{\extracolsep{\fill}}lcS[table-format=2.1]S[table-format=2.1]S[table-format=2.1]S[table-format=2.1]S[table-format=2.1]@{}}
\toprule
 & & \multicolumn{5}{c}{Number of options $K$} \\
\cmidrule(lr){3-7}
Method & Passes / q & {4} & {10} & {20} & {40} & {60} \\
\midrule
Baseline & 1 & 0.8 & 1.9 & 4.8 & 8.2 & 11.0 \\
\addlinespace[2pt]\nmdashrule\addlinespace[2pt]
Augmentation (2 ep.) & 1 & 0.6 & 2.5 & 6.2 & 8.6 & 11.8 \\
Augmentation (4 ep.) & 1 & 0.9 & 2.5 & 9.1 & 9.4 & 15.1 \\
Augmentation (4 ep.; fixed) & 1 & 1.1 & 5.4 & 10.8 & 16.4 & 19.5 \\
\addlinespace[2pt]\nmdashrule\addlinespace[2pt]
Random ensemble (avg, $\times 2$) & 2 & 0.5 & 1.3 & 2.6 & 4.8 & 6.8 \\
Random ensemble (avg, $\times 4$) & 4 & 0.4 & 0.9 & 1.9 & 3.3 & 4.1 \\
Random ensemble (avg, $\times 8$) & 8 & 0.1 & 0.7 & 1.3 & 2.1 & 2.8 \\
Cyclic ensemble (avg) & $K$ & 0.1 & 0.9 & 2.2 & 3.2 & 3.9 \\
\bottomrule
\end{NiceTabular*}
\vspace{-10pt}
\end{table}

%% file: table_released.tex
\begin{table}[t]
\vspace{-3pt}
\centering
\caption{Average accuracy (\avgacc{} $\uparrow$), worst-case accuracy (\worstacc{} $\uparrow$) and pairwise flip rate ($\widehat{\text{Flip}}$ $\downarrow$), in \%, averaged over the 3 out-of-domain sets and over all 7 sets. 
$^\dagger$ averaged over the subsets that the model supports; not compared. \textbf{Bold}: best in the size group.}
\label{tab:released}
\scriptsize
\setlength{\tabcolsep}{3pt}
\NiceMatrixOptions{custom-line={command=nmdashrule, tikz={dash pattern=on 1.5pt off 1.5pt}, total-width=0.3pt}}
\begin{NiceTabular*}{\columnwidth}{@{\extracolsep{\fill}}lcccccc@{}}
\toprule
 & \multicolumn{3}{c}{Mean (OOD)} & \multicolumn{3}{c}{Mean (all)} \\
\cmidrule(lr){2-4}\cmidrule(lr){5-7}
Model & \avgacc & \worstacc & $\widehat{\text{Flip}}$ & \avgacc & \worstacc & $\widehat{\text{Flip}}$ \\
\midrule
Baseline 1.7B & 59.5 & 47.1 & 17.7 & 67.9 & 57.5 & 12.7 \\
\Method{} 1.7B & 59.0 & \textbf{59.0} & \textbf{0.0} & 67.9 & \textbf{67.9} & \textbf{0.0} \\
Tacit 1.7B & 43.8$^\dagger$ & 26.1$^\dagger$ & 40.8$^\dagger$ & 46.8$^\dagger$ & 31.5$^\dagger$ & 33.3$^\dagger$ \\
d3-nano 2.2B & \textbf{63.5} & 53.7 & 12.5 & \textbf{70.2} & 61.7 & 9.1 \\
\addlinespace[2pt]\nmdashrule\addlinespace[2pt]
\Method{} 4B & 66.2 & \textbf{66.2} & \textbf{0.0} & 74.3 & \textbf{74.3} & \textbf{0.0} \\
Tacit 4B & 54.7$^\dagger$ & 37.1$^\dagger$ & 31.3$^\dagger$ & 56.6$^\dagger$ & 43.4$^\dagger$ & 23.6$^\dagger$ \\
Kev 4B & 63.8 & 43.8 & 24.1 & 68.9 & 50.2 & 17.7 \\
Jeva 4B & 57.2 & 57.2 & \textbf{0.0} & 64.1 & 64.1 & \textbf{0.0} \\
Intern-Decision 4B & 63.7 & 29.0 & 30.2 & 65.1$^\dagger$ & 39.3$^\dagger$ & 25.2$^\dagger$ \\
d3-mini 4.5B & \textbf{74.5} & \textbf{67.7} & 7.8 & \textbf{78.9} & \textbf{72.7} & 5.9 \\
\bottomrule
\end{NiceTabular*}
\vspace{-10pt}
\end{table}

%% file: 07_conclusion.tex
\section{Limitations}
Our controlled comparison uses one base model family and one training recipe, so the size of each effect may differ for other settings.
Our latency measurements cover a single GPU type and do not enable the encoding caching offered by separate encoding, whose benefit depends on specific cases.
Finally, tree attention assumes a standard attention mask; backbones with non-conventional attention layers, such as linear or hybrid attention, require their own implementation.

\section{Conclusion}
\label{sec:conclusion}
We studied how sensitive decision models are to the ordering of the options, and how to reduce this sensitivity.
Existing open- and closed-source decision models lose substantial accuracy in their worst-case order, and the loss grows with the number of options and compounds over the steps of a task.
Comparing eight methods on a common baseline model, we find that other methods either leave substantial sensitivity in place or remove it at a cost in accuracy or compute.
Among them, tree attention offers the most favorable trade-off, providing exact order invariance in a single forward pass without loss of accuracy when used in training, and at a cost of 1.1 points when applied to the trained baseline.
Overall, our results show that order invariance need not be traded off against accuracy or compute, and can instead be built into decision models by design.

%% file: 99_ack.tex
\section*{Acknowledgements}
This work was supported by the National Science Foundation (NSF) under grants OAC-2118310 and IIS-2530143, through the AI Research Institutes program (Award No. DMR-2433348), by a grant of access to OpenAI models through the ChatGPT for Academic Researchers program, by NewYork-Presbyterian for the NYP-Cornell Cardiovascular AI Collaboration, and by arXiv.
DG is supported by the Empire AI Postdoctoral Fellowship. 

%% file: R_related_work.tex
\section{Extended related work}
\label{app:related}
This appendix reviews related work on decision models and on order sensitivity, and the work behind the method families of \cref{sec:methods} in more detail.

\para{Decision models.}
Jev introduced one interface for several types of decision \citep{almeida2026jev}, and LLMs read out in the same way already act as decision models \citep{li2026llmasjev}.
Open decision models differ mainly in their read-out.
SemIf and JevK5 take option probabilities from the logits of answer tokens without decoding \citep{lee2026semif,jevk5}, Intern-Decision reads a distribution over permitted answer symbols at decision markers \citep{zhu2026intern}, Kev uses a pointer head over the contextualized positions of the options \citep{kev}, and Laya puts a decision head on a bidirectional encoder \citep{convai2026laya}.
A model that reads the options as one sequence inherits their order, whatever its read-out (\cref{fig:1}).
Among the remedies used for decision models, reflex aggregates predictions over several option orders \citep{raghavan2026reflex} and AnyJev adds a position prior to the cyclic ensemble \citep{anyjev2026}, both test-time ensembling, while Jeva encodes the options apart \citep{li2026jeva} (\cref{sec:methods}).
Released decision models also build in such remedies: Von and bev-decider read the options with tree attention \citep{von,biswas2026bevdecider}, and Runoff filters the options before a cyclic ensemble \citep{huang2026runoff}.

\para{Order sensitivity beyond decision models.}
Order sensitivity has been studied well beyond decision models.
In few-shot prompting, the order of demonstrations can move a model from near state of the art to chance \citep{lu2022fantastically}, which motivated calibration \citep{zhao2021calibrate,fei2023label,zhou2024batch} and order-agnostic aggregation of demonstrations \citep{zhang2024batchicl}.
LLM judges prefer one answer position \citep{zheng2023judging,wang2024fair,shi2025judging}, and long-context models neglect the middle of their input \citep{liu2024lost}; remedies there recalibrate or rescale positional attention \citep{hsieh2024found,yu2025mitigate}, which needs access to the model's internals, as do pruning methods that remove biased components \citep{zhou2024unibias,choi2025mitigating}.
More recent work trains for consistency across permutations with reinforcement learning \citep{zheng2026pagrpo,chu2026dgao}, or notes that reasoning at length reduces position bias \citep{li2026comprehensive}, though not always \citep{wang2026morethinking}; both rely on generating text and so do not apply to one-pass decision models.

\para{Post-hoc calibration.}
Typically, methods in this family rescale and shift the log-probability of each position with parameters shared by all items, a form known as vector scaling \citep{guo2017calibration,zhao2021calibrate}; the methods differ in how these parameters are set.
Label-free methods estimate the model's average preference for each position and divide it out; the estimate comes from an input with no content \citep{zhao2021calibrate,lv2026emptiness} or random in-domain words \citep{fei2023label}, from the mean output over a batch of unlabelled items \citep{zhou2024batch}, or from a few items answered in several orders \citep{zheng2024selectors}.
Supervised methods fit the parameters on labelled items from the target distribution \citep{guo2017calibration}.
PriDe \citep{zheng2024selectors} estimates the prior from a few items answered in several orders, so that their content cancels and only the preference for positions remains.
Its original use was to remove the model prior tied to specific option ID tokens (e.g., A/B/C); \cref{sec:methods} applies it to positions.
Per-position temperature scaling is a form of vector scaling \citep{guo2017calibration} with one scale per position.
A single shared temperature cannot change the argmax, so any effect on the answer comes from the differences between positions.

\para{Test-time ensembling.}
Test-time ensembling asks the model the same question under a set of option orders and aggregates the answers, at one forward pass per order; the passes are independent and can run in parallel.
The answers are combined by averaging the option probabilities (\emph{avg}) \citep{zheng2024selectors}, by majority vote (\emph{vote}) \citep{pezeshkpour2024order}, or by taking their geometric mean \citep{anyjev2026}.
Averaging over all $K!$ orders makes the answer exactly order-invariant \citep{izacard2023atlas,zheng2024selectors}; a smaller set of orders saves cost, but keeps the answer invariant only if the set does not depend on the given order, that is, if it is built from a canonical listing of the options \citep{puny2022frame,kaba2023canonicalization}.
The random-permutation ensemble \citep{pezeshkpour2024order,hou2024rankers,tang2024found} uses $m$ random orders, whose answers are aggregated; as $m$ grows it approaches the average over all $K!$ orders, but for any smaller $m$ another draw can give another answer.
The cyclic-permutation ensemble \citep{izacard2023atlas,zheng2024selectors} uses the $K$ cyclic permutations of a start order, which form a Latin square: the smallest set in which each option takes each position once.
Besides avg and vote, AnyJev \citep{anyjev2026} combines it with the \emph{L0} reducer, which estimates the model's prior over positions from the evaluation set and divides it out of each answer before taking the geometric mean (Appendix~\ref{app:methods}).
The filtered cyclic-permutation ensemble first shortlists the options with a filter and then ensembles over the shortlist only.
Shortlisting before the final choice is common across domains such as document re-ranking \citep{pradeep2023rankvicuna,tang2024found}, web agents \citep{deng2023mind2web}, multiple-choice question answering \citep{ma2023poe,zhu2025optionid}, and tool selection \citep{blankenstein2026biasbusters}.
If the filter scores each option alone and fixes the order of the shortlist, the answer is also order-invariant by construction.
Runoff \citep{huang2026runoff} was among the first to formalize this idea for decision models.

\para{Training-time augmentation.}
Training-time augmentation shows each training question under several option orders, as shuffled copies added to the training set or as a fresh order at each step, so that the model learns not to rely on position.
It is used for listwise rerankers \citep{pradeep2023rankvicuna,qiao2026listwise}, LLM judges \citep{zhu2025judgelm,whitehouse2025j1}, and decision models \citep{li2026llmasjev}.
Stronger variants add a loss term that ties the answers across orders \citep{frohmann2026equal,xiang2024addressing} or train against the worst-case orders \citep{chen2025pearl}.

\para{Permutation invariance.}
Formally, averaging a sequence model over all orders yields a permutation-invariant function \citep{murphy2019janossy}, which test-time ensembling and training-time augmentation approximate, while set architectures obtain invariance by construction \citep{zaheer2017deepsets,lee2019settransformer}.
For decision models, this design has already been implemented. In Von 1.2, an option's tokens attend only to the shared premise and to that option's own tokens, and every option's position ids restart at the end of the premise, which removed all answer flips on its order test \citep{von}. bev-decider starts every option at the same position id and reads the options in parallel \citep{biswas2026bevdecider}. \citet{amit2026candidate} trains Qwen3-based decision models with candidate-independent block-causal attention, in which options cannot attend to one another and share their start position. Our tree attention follows these implementations.
In language models, attention masks with shared position ids make a set of inputs order-invariant, without fine-tuning \citep{mcilroy2024order} or with it \citep{brown2025finetuning,egressy2025set}; related inference-time variants re-assign positions by attention \citep{wang2024pine} or rotate them \citep{yoon2025rotor}. The same mask structure, with independent blocks that share a prefix, serves other goals: scaling in-context learning \citep{hao2022structured,ratner2023pcw}, parallel encoding of retrieved contexts \citep{yang2025ape}, verifying candidate continuations in speculative decoding \citep{miao2024specinfer,cai2024medusa}, and listwise reranking of passages and items \citep{schlatt2025setencoder,bito2026invarirank}.

%% file: A_metric_tables.tex
\section{Full metric tables}
\label{app:tables}
This appendix reports every metric for every method arm, open-weight model and hosted API on every evaluation set; the tables and figures in the main text show subsets of these.

\input{appA_notes.tex}

\para{All metrics.} Average and worst-case accuracy, the accuracy drop and the pairwise flip rate of every method arm, open-weight model and hosted API on every set: \cref{tab:app-systems}.
\input{appA_systems.tex}

\para{Scaling with the number of options.} \cref{tab:app-scaling-1,tab:app-scaling-2}.
\input{appA_scaling_1.tex}
\input{appA_scaling_2.tex}

\para{Mind2Web task success.} \cref{tab:app-m2w-tasks}.
\input{appA_m2w_tasks.tex}

\para{How tight the Latin-square worst is.} \cref{tab:app-all-orders}.
\input{appA_all_orders.tex}

\clearpage

%% file: appA_notes.tex
\para{How to read these tables.} Cells: mean $\pm$ standard deviation over 3 training seeds for the arms trained on Qwen3-1.7B-Base (the baseline model, its post-hoc calibrations and test-time ensembles, augmentation, separate encoding with its Jeva's-head ablation and Jeva's-implementation control, and \Method{} 1.7B trained or at test time). Order-invariant arms: \method{}, the filtered cyclic-permutation ensemble and Jeva's implementation have seed 0 in every Latin-square order and seeds 1 and 2 in the original order only, so their pairwise flip rate, \gap{} and Var.\ are seed 0's (no $\pm$); separate encoding and its Jeva's-head ablation have all three seeds in the original order only, so \worstacc{} $=$ \avgacc{} and \gap{} and the flip rate are 0 by construction; \method{} at test time has all three seeds in every Latin-square order. Var.\ columns show the mean; their captions give the largest standard deviation. Other rows are one checkpoint or API: one value. Every method is evaluated $K$ times per question (the permutations $\Pi_n$, \cref{sec:defs}); average accuracy \avgacc{} = mean over the evaluations, worst-case accuracy \worstacc{} = right in all of them, pairwise flip rate = chance that two evaluations give different answers, accuracy drop \gap{} = \avgacc{} $-$ \worstacc{}, in points. Var.: order variance $\times 100$, the mean over questions of $p(1{-}p)$ with $p$ the share of the question's evaluations answered right (0 = order-invariant, at most 25). Rows: class, then method, then size. $^\dagger$ not trained on Qwen3-1.7B-Base with our data (the same-base rows are listed above). $^{a}$ PriDe: prior estimated label-free on 5\% of the items, scored on the other 95\% (mean of 5 draws); temperature fitted on gold labels, cross-fitted over 2 folds. $^{b}$ $K$ evaluations of each question (fewer where $K!$ is too small: 3 at $K{=}4$, $m{=}8$), each answered from $m$ fresh random orders, $m$ passes per evaluation; \worstacc: right in all evaluations, \avgacc: mean over them, flip rate: two evaluations disagree; no original-order value (--). $^{c}$ $K$ evaluations of each question, one per Latin-square order (the orders of the one-pass rows), each answered from that order's $K$ cyclic permutations (avg: mean probability; L0: geometric mean after dividing out the position prior; vote: plurality of the $K$ picks; L0, vote: the same vote on the prior-corrected cyclic permutations), $K$ passes per evaluation; \worstacc, \avgacc{} and flip rate as in $^{b}$; not run above $K{=}60$ (n/a). L0's per-row prior is estimated from few questions per option on MASSIVE at $K{=}40,60$ and CLINC150 at $K{=}50$ (3--5), which lowers its worst-case accuracy there; a prior pooled over the rows recovers the avg rule's accuracy. The rows shown use the per-row prior. Tree attention (test-time): the baseline's weights scored with \method{}'s encoding, not retrained. Separate encoding: our prompt and a cosine head; its ablation swaps in Jeva's scoring head; Jeva's implementation, with its own prompt and head, is the control. $^{f}$ Filter, then ensemble; instance tested: Runoff \citep{huang2026runoff}, here with the baseline model for both stages: one pass asks yes/no for every option, then the shortlist $C$ is answered in each of its $|C|$ cyclic permutations and averaged; order-invariant by construction. Mean passes per question (seed 0): 4.8 on MMLU, 10.4 on MMLU-Pro, 98.1 on CLINC150 (151 options). n/a: impossible by design (Tacit accepts at most 26 options, Intern-Decision at most 62; $K{=}2$ has only two orders, so no random-permutation ensemble). $^\star$ in-domain (training splits are in our training data). Mind2Web: 300 tasks, 2118 steps, $K=10$ candidates per step; per step, every step is one question; per task, a task counts as right only if all its steps are (one task = 0.33 points), with the same order at every step, and random-permutation ensembles are the mean over 5 random assignments of orders to evaluations. Each system is scored on all items it answered.

%% file: appA_systems.tex
\begin{sidewaystable*}
\definecolor{invrow}{RGB}{222,235,250}\definecolor{lowcell}{RGB}{252,222,222}
\centering
\caption{Every method arm, open-weight model and hosted API: average accuracy (\avgacc{} $\uparrow$), worst-case accuracy (\worstacc{} $\uparrow$), accuracy drop (\gap{} $\downarrow$) and pairwise flip rate (Flip $\downarrow$), in \%. Mean (OOD): average over the 3 out-of-domain sets; Mean (all): over all 7; n/a if a set is missing or not supported. {\setlength{\fboxsep}{0.8pt}\colorbox{invrow}{Blue}}: order-invariant. {\setlength{\fboxsep}{0.8pt}\colorbox{lowcell}{Red}}: significantly worse than the baseline model (95\% paired interval). Method arms: mean over $n{=}3$ training seeds (order-invariant arms: seed 0's \gap{} and flip rate). Hosted APIs: each on the items it answered. $^\star$ out-of-domain. Options per question $K$: MMLU 4, MMLU-Pro 10, MASSIVE 60, ARC 4, BANKING77 77, CLINC150 151, Mind2Web 10. Other marks and details: \cref{app:tables}.}
\label{tab:app-systems}
\tiny
\setlength{\tabcolsep}{2pt}
\NiceMatrixOptions{custom-line={command=nmdashrule, tikz={dash pattern=on 1.5pt off 1.5pt}, total-width=0.3pt}}
\begin{NiceTabular}{@{}l@{\hspace{9pt}}w{c}{11pt}w{c}{11pt}w{c}{11pt}w{c}{11pt}Xw{c}{11pt}w{c}{11pt}w{c}{11pt}w{c}{11pt}Xw{c}{11pt}w{c}{11pt}w{c}{11pt}w{c}{11pt}Xw{c}{11pt}w{c}{11pt}w{c}{11pt}w{c}{11pt}Xw{c}{11pt}w{c}{11pt}w{c}{11pt}w{c}{11pt}Xw{c}{11pt}w{c}{11pt}w{c}{11pt}w{c}{11pt}Xw{c}{11pt}w{c}{11pt}w{c}{11pt}w{c}{11pt}Xw{c}{11pt}w{c}{11pt}w{c}{11pt}w{c}{11pt}Xw{c}{11pt}w{c}{11pt}w{c}{11pt}w{c}{11pt}@{}}[width=\linewidth, create-extra-nodes]
\toprule
 & \multicolumn{4}{c}{\makebox[0pt]{MMLU$^\star$}} & & \multicolumn{4}{c}{\makebox[0pt]{MMLU-Pro$^\star$}} & & \multicolumn{4}{c}{\makebox[0pt]{MASSIVE$^\star$}} & & \multicolumn{4}{c}{\makebox[0pt]{ARC}} & & \multicolumn{4}{c}{\makebox[0pt]{BANKING77}} & & \multicolumn{4}{c}{\makebox[0pt]{CLINC150}} & & \multicolumn{4}{c}{\makebox[0pt]{Mind2Web}} & & \multicolumn{4}{c}{\makebox[0pt]{Mean (OOD)}} & & \multicolumn{4}{c}{\makebox[0pt]{Mean (all)}} \\
\cmidrule(lr){2-5}\cmidrule(lr){7-10}\cmidrule(lr){12-15}\cmidrule(lr){17-20}\cmidrule(lr){22-25}\cmidrule(lr){27-30}\cmidrule(lr){32-35}\cmidrule(lr){37-40}\cmidrule(lr){42-45}
System & {\makebox[0pt]{\tiny\hspace{-6pt}\avgacc}} & {\makebox[0pt]{\tiny\worstacc}} & {\makebox[0pt]{\tiny\gap}} & {\makebox[0pt]{\tiny{Flip}}} & & {\makebox[0pt]{\tiny\hspace{-6pt}\avgacc}} & {\makebox[0pt]{\tiny\worstacc}} & {\makebox[0pt]{\tiny\gap}} & {\makebox[0pt]{\tiny{Flip}}} & & {\makebox[0pt]{\tiny\hspace{-6pt}\avgacc}} & {\makebox[0pt]{\tiny\worstacc}} & {\makebox[0pt]{\tiny\gap}} & {\makebox[0pt]{\tiny{Flip}}} & & {\makebox[0pt]{\tiny\hspace{-6pt}\avgacc}} & {\makebox[0pt]{\tiny\worstacc}} & {\makebox[0pt]{\tiny\gap}} & {\makebox[0pt]{\tiny{Flip}}} & & {\makebox[0pt]{\tiny\hspace{-6pt}\avgacc}} & {\makebox[0pt]{\tiny\worstacc}} & {\makebox[0pt]{\tiny\gap}} & {\makebox[0pt]{\tiny{Flip}}} & & {\makebox[0pt]{\tiny\hspace{-6pt}\avgacc}} & {\makebox[0pt]{\tiny\worstacc}} & {\makebox[0pt]{\tiny\gap}} & {\makebox[0pt]{\tiny{Flip}}} & & {\makebox[0pt]{\tiny\hspace{-6pt}\avgacc}} & {\makebox[0pt]{\tiny\worstacc}} & {\makebox[0pt]{\tiny\gap}} & {\makebox[0pt]{\tiny{Flip}}} & & {\makebox[0pt]{\tiny\hspace{-6pt}\avgacc}} & {\makebox[0pt]{\tiny\worstacc}} & {\makebox[0pt]{\tiny\gap}} & {\makebox[0pt]{\tiny{Flip}}} & & {\makebox[0pt]{\tiny\hspace{-6pt}\avgacc}} & {\makebox[0pt]{\tiny\worstacc}} & {\makebox[0pt]{\tiny\gap}} & {\makebox[0pt]{\tiny{Flip}}} \\
\midrule
Baseline model & 62.4 & 53.3 & 9.1 & 14.6 & & 38.7 & 21.6 & 17.1 & 32.2 & & 77.5 & 66.5 & 11.0 & 6.3 & & 85.0 & 80.1 & 4.9 & 6.4 & & 88.9 & 81.0 & 7.9 & 4.4 & & 89.6 & 78.8 & 10.7 & 4.3 & & 33.6 & 21.3 & 12.3 & 20.4 & & 59.5 & 47.1 & 12.4 & 17.7 & & 67.9 & 57.5 & 10.4 & 12.7 \\
\addlinespace[2pt]\nmdashrule\addlinespace[2pt]
\multicolumn{45}{l}{\textit{Post-hoc calibration}} \\
PriDe (5\%) & 62.6 & 53.4 & 9.3 & 14.7 & & 38.7 & 20.9 & 17.8 & 32.6 & & 77.5 & 62.3 & 15.2 & 7.2 & & 85.0 & 80.0 & 5.0 & 6.6 & & 88.8 & 77.4 & 11.4 & 4.6 & & 89.2 & 72.3 & 16.9 & 5.0 & & 33.6 & 21.7 & 11.9 & 20.2 & & 59.6 & 45.5 & 14.1 & 18.1 & & 67.9 & 55.4 & 12.5 & 13.0 \\
Per-position temp. & 62.5 & 53.4 & 9.1 & 14.5 & & 38.8 & 21.5 & 17.3 & 32.2 & & 77.5 & 66.3 & 11.2 & 6.3 & & 85.0 & 80.0 & 5.0 & 6.4 & & 88.9 & 81.0 & 7.9 & 4.3 & & 89.6 & 78.8 & 10.7 & 4.3 & & 33.5 & 21.5 & 12.0 & 20.0 & & 59.6 & 47.1 & 12.5 & 17.7 & & 68.0 & 57.5 & 10.4 & 12.6 \\
\addlinespace[2pt]\nmdashrule\addlinespace[2pt]
\multicolumn{45}{l}{\textit{Training-time augmentation}} \\
Augmentation (2 ep.) & 61.5 & 53.3 & 8.2 & 12.9 & & 37.8 & 22.4 & 15.5 & 30.0 & & 76.9 & 65.2 & 11.8 & 7.1 & & 84.9 & 80.0 & 4.9 & 6.2 & & 91.5 & 84.8 & 6.7 & 3.1 & & 92.0 & 81.2 & 10.8 & 3.6 & & 35.6 & 24.7 & 10.9 & 18.1 & & 58.7 & 46.9 & 11.8 & 16.7 & & 68.6 & 58.8 & 9.8 & 11.6 \\
Augmentation (4 ep.) & 61.0 & 53.4 & 7.6 & 11.9 & & 35.1 & 21.8 & 13.3 & 27.8 & & 75.3 & 60.2 & 15.1 & 8.7 & & 84.1 & 80.3 & 3.9 & 5.3 & & 91.8 & 84.4 & 7.3 & 3.6 & & 91.8 & 82.2 & 9.6 & 3.3 & & 34.5 & 24.9 & 9.6 & 16.7 & & 57.1 & 45.1 & 12.0 & 16.1 & & 67.7 & 58.2 & 9.5 & 11.0 \\
Augmentation (4 ep.; fixed) & 60.6 & 48.7 & 11.9 & 18.5 & & 36.0 & 17.1 & 18.9 & 39.4 & & 74.0 & 54.5 & 19.5 & 12.5 & & 83.9 & 76.9 & 7.1 & 9.0 & & 90.9 & 79.0 & 11.9 & 5.7 & & 91.3 & 75.2 & 16.1 & 5.7 & & 33.1 & 19.4 & 13.7 & 24.6 & & 56.9 & 40.1 & 16.8 & 23.4 & & 67.1 & 53.0 & 14.2 & 16.5 \\
\addlinespace[2pt]\nmdashrule\addlinespace[2pt]
\multicolumn{45}{l}{\textit{Test-time ensembling}} \\
Random ensemble ($\times 2$) & 62.8 & 56.9 & 5.9 & 9.0 & & 39.5 & 26.3 & 13.3 & 23.9 & & 77.8 & 71.0 & 6.8 & 4.3 & & 85.4 & 82.2 & 3.2 & 4.1 & & 89.0 & 84.5 & 4.5 & 3.1 & & 89.8 & 83.1 & 6.7 & 3.0 & & 33.9 & 25.3 & 8.5 & 14.2 & & 60.0 & 51.4 & 8.6 & 12.4 & & 68.3 & 61.3 & 7.0 & 8.8 \\
Random ensemble ($\times 4$) & 63.0 & 58.7 & 4.3 & 6.6 & & 40.0 & 29.8 & 10.3 & 18.2 & & 77.9 & 73.8 & 4.1 & 3.0 & & 85.6 & 83.4 & 2.2 & 2.9 & & 89.1 & 85.8 & 3.3 & 2.3 & & 90.0 & 85.2 & 4.7 & 2.2 & & 34.1 & 27.9 & 6.2 & 10.3 & & 60.3 & 54.1 & 6.2 & 9.3 & & 68.5 & 63.5 & 5.0 & 6.5 \\
Random ensemble ($\times 8$) & 63.2 & 60.5 & 2.7 & 4.9 & & 40.4 & 32.6 & 7.8 & 13.5 & & 77.9 & 75.1 & 2.8 & 2.1 & & 85.8 & 84.5 & 1.3 & 2.1 & & 89.1 & 86.8 & 2.3 & 1.6 & & 90.1 & 86.7 & 3.3 & 1.5 & & 34.2 & 29.8 & 4.4 & 7.4 & & 60.5 & 56.1 & 4.4 & 6.8 & & 68.7 & 65.1 & 3.5 & 4.7 \\
Cyclic ensemble (avg) & 63.2 & 60.7 & 2.5 & 3.7 & & 40.4 & 31.7 & 8.7 & 14.6 & & 78.0 & 74.2 & 3.9 & 2.6 & & 85.6 & 84.1 & 1.4 & 1.9 & & {\textit{n/a}} & {\textit{n/a}} & {\textit{n/a}} & {\textit{n/a}} & & {\textit{n/a}} & {\textit{n/a}} & {\textit{n/a}} & {\textit{n/a}} & & 34.1 & 28.3 & 5.8 & 9.5 & & 60.5 & 55.5 & 5.0 & 7.0 & & {\textit{n/a}} & {\textit{n/a}} & {\textit{n/a}} & {\textit{n/a}} \\
Cyclic ensemble (L0) & 63.1 & 60.5 & 2.6 & 4.1 & & 40.2 & 31.4 & 8.8 & 14.9 & & 77.7 & 66.0 & 11.7 & 5.9 & & 85.6 & 83.8 & 1.8 & 2.1 & & {\textit{n/a}} & {\textit{n/a}} & {\textit{n/a}} & {\textit{n/a}} & & {\textit{n/a}} & {\textit{n/a}} & {\textit{n/a}} & {\textit{n/a}} & & 34.1 & 28.3 & 5.8 & 9.7 & & 60.4 & 52.6 & 7.7 & 8.3 & & {\textit{n/a}} & {\textit{n/a}} & {\textit{n/a}} & {\textit{n/a}} \\
Filtered ensemble & 63.2 & 63.2 & 0.0 & 0.0 & & 39.8 & 39.8 & 0.0 & 0.0 & & 77.8 & 77.8 & 0.0 & 0.0 & & 85.4 & 85.4 & 0.0 & 0.0 & & 89.4 & 89.4 & 0.0 & 0.0 & & 89.8 & 89.8 & 0.0 & 0.0 & & 34.0 & 34.0 & 0.0 & 0.0 & & 60.3 & 60.3 & 0.0 & 0.0 & & 68.5 & 68.5 & 0.0 & 0.0 \\
\addlinespace[2pt]\nmdashrule\addlinespace[2pt]
\multicolumn{45}{l}{\textit{Architectural designs}} \\
Separate encoding & 51.4 & 51.4 & 0.0 & 0.0 & & 28.8 & 28.8 & 0.0 & 0.0 & & 66.3 & 66.3 & 0.0 & 0.0 & & 71.6 & 71.6 & 0.0 & 0.0 & & 90.6 & 90.6 & 0.0 & 0.0 & & 85.5 & 85.5 & 0.0 & 0.0 & & 29.4 & 29.4 & 0.0 & 0.0 & & 48.8 & 48.8 & 0.0 & 0.0 & & 60.5 & 60.5 & 0.0 & 0.0 \\
Sep. encoding (Jeva head) & 48.9 & 48.9 & 0.0 & 0.0 & & 27.4 & 27.4 & 0.0 & 0.0 & & 59.7 & 59.7 & 0.0 & 0.0 & & 68.7 & 68.7 & 0.0 & 0.0 & & 90.9 & 90.9 & 0.0 & 0.0 & & 87.0 & 87.0 & 0.0 & 0.0 & & 25.2 & 25.2 & 0.0 & 0.0 & & 45.3 & 45.3 & 0.0 & 0.0 & & 58.3 & 58.3 & 0.0 & 0.0 \\
Sep. encoding (Jeva impl.) & 48.7 & 48.7 & 0.0 & 0.0 & & 26.9 & 26.9 & 0.0 & 0.0 & & 51.5 & 51.5 & 0.0 & 0.0 & & 70.3 & 70.3 & 0.0 & 0.0 & & 91.2 & 91.2 & 0.0 & 0.0 & & 84.7 & 84.7 & 0.0 & 0.0 & & 23.3 & 23.3 & 0.0 & 0.0 & & 42.4 & 42.4 & 0.0 & 0.0 & & 56.7 & 56.7 & 0.0 & 0.0 \\
\Method{} (test-time) & 61.4 & 61.4 & 0.0 & 0.0 & & 37.4 & 37.4 & 0.0 & 0.0 & & 77.2 & 77.2 & 0.0 & 0.0 & & 84.6 & 84.6 & 0.0 & 0.0 & & 86.9 & 86.9 & 0.0 & 0.0 & & 87.3 & 87.3 & 0.0 & 0.0 & & 32.7 & 32.7 & 0.0 & 0.0 & & 58.6 & 58.6 & 0.0 & 0.0 & & 66.8 & 66.8 & 0.0 & 0.0 \\
\Method{} & 62.4 & 62.4 & 0.0 & 0.0 & & 37.9 & 37.9 & 0.0 & 0.0 & & 76.7 & 76.7 & 0.0 & 0.0 & & 85.2 & 85.2 & 0.0 & 0.0 & & 89.6 & 89.6 & 0.0 & 0.0 & & 90.3 & 90.3 & 0.0 & 0.0 & & 33.0 & 33.0 & 0.0 & 0.1 & & 59.0 & 59.0 & 0.0 & 0.0 & & 67.9 & 67.9 & 0.0 & 0.0 \\
\midrule
\multicolumn{45}{l}{\textit{Open-weight models (matching parameter size)}} \\
Tacit 1.7B$^\dagger$ & 57.2 & 41.3 & 15.9 & 26.2 & & 30.3 & 10.9 & 19.4 & 55.4 & & {\textit{n/a}} & {\textit{n/a}} & {\textit{n/a}} & {\textit{n/a}} & & 77.2 & 66.3 & 10.9 & 13.7 & & {\textit{n/a}} & {\textit{n/a}} & {\textit{n/a}} & {\textit{n/a}} & & {\textit{n/a}} & {\textit{n/a}} & {\textit{n/a}} & {\textit{n/a}} & & 22.6 & 7.7 & 14.9 & 37.9 & & {\textit{n/a}} & {\textit{n/a}} & {\textit{n/a}} & {\textit{n/a}} & & {\textit{n/a}} & {\textit{n/a}} & {\textit{n/a}} & {\textit{n/a}} \\
d3-nano 2.2B$^\dagger$ & 66.3 & 59.5 & 6.7 & 10.8 & & 39.9 & 26.6 & 13.3 & 23.2 & & 84.5 & 75.0 & 9.5 & 3.7 & & 88.9 & 85.8 & 3.1 & 3.6 & & 89.7 & 81.7 & 8.0 & 4.3 & & 92.1 & 81.0 & 11.1 & 2.7 & & 30.3 & 22.5 & 7.9 & 15.6 & & 63.5 & 53.7 & 9.8 & 12.5 & & 70.2 & 61.7 & 8.5 & 9.1 \\
Intern-Decision 4B$^\dagger$ & 73.7 & 61.4 & 12.3 & 18.6 & & 43.6 & 20.2 & 23.4 & 49.1 & & 73.7 & 5.5 & 68.2 & 22.9 & & 93.2 & 87.7 & 5.5 & 6.0 & & {\textit{n/a}} & {\textit{n/a}} & {\textit{n/a}} & {\textit{n/a}} & & {\textit{n/a}} & {\textit{n/a}} & {\textit{n/a}} & {\textit{n/a}} & & 41.5 & 21.9 & 19.6 & 29.2 & & 63.7 & 29.0 & 34.6 & 30.2 & & {\textit{n/a}} & {\textit{n/a}} & {\textit{n/a}} & {\textit{n/a}} \\
Jeva 4B$^\dagger$ & 52.9 & 52.9 & 0.0 & 0.0 & & 29.7 & 29.7 & 0.0 & 0.0 & & 89.0 & 89.0 & 0.0 & 0.0 & & 77.7 & 77.7 & 0.0 & 0.0 & & 92.0 & 92.0 & 0.0 & 0.0 & & 77.0 & 77.0 & 0.0 & 0.0 & & 30.3 & 30.3 & 0.0 & 0.0 & & 57.2 & 57.2 & 0.0 & 0.0 & & 64.1 & 64.1 & 0.0 & 0.0 \\
Kev 4B$^\dagger$ & 73.5 & 62.4 & 11.0 & 15.9 & & 49.1 & 28.0 & 21.2 & 38.4 & & 68.7 & 41.0 & 27.7 & 17.9 & & 93.2 & 89.1 & 4.1 & 4.6 & & 85.4 & 70.3 & 15.1 & 7.5 & & 75.8 & 38.0 & 37.8 & 16.3 & & 36.8 & 22.6 & 14.2 & 23.4 & & 63.8 & 43.8 & 20.0 & 24.1 & & 68.9 & 50.2 & 18.7 & 17.7 \\
Tacit 4B$^\dagger$ & 69.2 & 55.8 & 13.4 & 18.8 & & 40.1 & 18.3 & 21.8 & 43.8 & & {\textit{n/a}} & {\textit{n/a}} & {\textit{n/a}} & {\textit{n/a}} & & 88.7 & 83.2 & 5.6 & 6.9 & & {\textit{n/a}} & {\textit{n/a}} & {\textit{n/a}} & {\textit{n/a}} & & {\textit{n/a}} & {\textit{n/a}} & {\textit{n/a}} & {\textit{n/a}} & & 28.3 & 16.2 & 12.2 & 24.9 & & {\textit{n/a}} & {\textit{n/a}} & {\textit{n/a}} & {\textit{n/a}} & & {\textit{n/a}} & {\textit{n/a}} & {\textit{n/a}} & {\textit{n/a}} \\
d3-mini 4.5B$^\dagger$ & 80.5 & 76.7 & 3.8 & 5.0 & & 56.5 & 45.4 & 11.1 & 16.2 & & 86.6 & 81.0 & 5.6 & 2.3 & & 95.5 & 94.3 & 1.2 & 1.4 & & 91.7 & 84.0 & 7.7 & 2.7 & & 94.9 & 88.5 & 6.4 & 1.9 & & 46.8 & 39.0 & 7.9 & 11.8 & & 74.5 & 67.7 & 6.8 & 7.8 & & 78.9 & 72.7 & 6.2 & 5.9 \\
Elephant-1 9B$^\dagger$ & 83.0 & 76.3 & 6.7 & 8.5 & & 67.1 & 55.5 & 11.6 & 14.1 & & 90.1 & 86.5 & 3.6 & 1.6 & & 96.1 & 94.3 & 1.8 & 1.9 & & 90.4 & 77.0 & 13.4 & 4.4 & & 95.2 & 88.5 & 6.7 & 2.0 & & 55.1 & 49.2 & 5.9 & 7.2 & & 80.1 & 72.8 & 7.3 & 8.0 & & 82.4 & 75.3 & 7.1 & 5.7 \\
Mammoth-1 27B$^\dagger$ & 89.3 & 85.5 & 3.8 & 4.6 & & 77.3 & 66.9 & 10.4 & 11.4 & & 90.9 & 88.0 & 2.9 & 2.2 & & 97.5 & 96.5 & 1.0 & 1.1 & & 92.5 & 85.3 & 7.2 & 3.3 & & 95.6 & 93.0 & 2.6 & 1.1 & & 64.8 & 60.3 & 4.4 & 5.1 & & 85.9 & 80.2 & 5.7 & 6.1 & & 86.8 & 82.2 & 4.6 & 4.1 \\
\addlinespace[2pt]\nmdashrule\addlinespace[2pt]
\multicolumn{45}{l}{\textit{Hosted APIs}} \\
Jev API & 91.8 & 90.1 & 1.7 & 2.1 & & 81.3 & 69.8 & 11.5 & 9.6 & & 80.3 & 68.0 & 12.3 & 8.5 & & 97.8 & 97.4 & 0.4 & 0.6 & & 80.6 & 67.7 & 12.9 & 7.0 & & 88.3 & 71.4 & 17.0 & 5.4 & & 54.5 & 46.8 & 7.6 & 8.8 & & 84.5 & 76.0 & 8.5 & 6.7 & & 82.1 & 73.0 & 9.1 & 6.0 \\
OpenAI API & 85.4 & 77.7 & 7.7 & 10.0 & & 56.2 & 35.1 & 21.1 & 37.9 & & 81.6 & 48.1 & 33.5 & 14.1 & & 96.4 & 93.9 & 2.5 & 2.7 & & 76.8 & 46.3 & 30.5 & 15.0 & & 85.6 & 37.8 & 47.8 & 13.5 & & 51.9 & 34.6 & 17.3 & 21.3 & & 74.4 & 53.6 & 20.8 & 20.7 & & 76.3 & 53.4 & 22.9 & 16.3 \\
\bottomrule
\CodeAfter
\providecommand{\cellbox}[2]{\tikz\fill[#1, blend mode=multiply] ([xshift=-2pt]#2-medium.north west) rectangle ([xshift=2pt]#2-medium.south east);}
\cellbox{lowcell}{5-8}
\cellbox{lowcell}{5-9}
\cellbox{lowcell}{5-10}
\cellbox{lowcell}{5-13}
\cellbox{lowcell}{5-14}
\cellbox{lowcell}{5-15}
\cellbox{lowcell}{5-20}
\cellbox{lowcell}{5-23}
\cellbox{lowcell}{5-24}
\cellbox{lowcell}{5-28}
\cellbox{lowcell}{5-29}
\cellbox{lowcell}{5-38}
\cellbox{lowcell}{5-39}
\cellbox{lowcell}{5-40}
\cellbox{lowcell}{5-43}
\cellbox{lowcell}{5-44}
\cellbox{lowcell}{5-45}
\cellbox{lowcell}{6-32}
\cellbox{lowcell}{8-2}
\cellbox{lowcell}{8-7}
\cellbox{lowcell}{8-37}
\cellbox{lowcell}{9-2}
\cellbox{lowcell}{9-7}
\cellbox{lowcell}{9-12}
\cellbox{lowcell}{9-13}
\cellbox{lowcell}{9-14}
\cellbox{lowcell}{9-15}
\cellbox{lowcell}{9-37}
\cellbox{lowcell}{9-38}
\cellbox{lowcell}{10-2}
\cellbox{lowcell}{10-3}
\cellbox{lowcell}{10-4}
\cellbox{lowcell}{10-5}
\cellbox{lowcell}{10-7}
\cellbox{lowcell}{10-8}
\cellbox{lowcell}{10-9}
\cellbox{lowcell}{10-10}
\cellbox{lowcell}{10-12}
\cellbox{lowcell}{10-13}
\cellbox{lowcell}{10-14}
\cellbox{lowcell}{10-15}
\cellbox{lowcell}{10-17}
\cellbox{lowcell}{10-18}
\cellbox{lowcell}{10-19}
\cellbox{lowcell}{10-20}
\cellbox{lowcell}{10-24}
\cellbox{lowcell}{10-25}
\cellbox{lowcell}{10-28}
\cellbox{lowcell}{10-29}
\cellbox{lowcell}{10-30}
\cellbox{lowcell}{10-33}
\cellbox{lowcell}{10-34}
\cellbox{lowcell}{10-35}
\cellbox{lowcell}{10-37}
\cellbox{lowcell}{10-38}
\cellbox{lowcell}{10-39}
\cellbox{lowcell}{10-40}
\cellbox{lowcell}{10-42}
\cellbox{lowcell}{10-43}
\cellbox{lowcell}{10-44}
\cellbox{lowcell}{10-45}
\cellbox{lowcell}{19-2}
\cellbox{lowcell}{19-7}
\cellbox{lowcell}{19-12}
\cellbox{lowcell}{19-17}
\cellbox{lowcell}{19-18}
\cellbox{lowcell}{19-27}
\cellbox{lowcell}{19-32}
\cellbox{lowcell}{19-37}
\cellbox{lowcell}{19-42}
\cellbox{lowcell}{20-2}
\cellbox{lowcell}{20-3}
\cellbox{lowcell}{20-7}
\cellbox{lowcell}{20-12}
\cellbox{lowcell}{20-13}
\cellbox{lowcell}{20-17}
\cellbox{lowcell}{20-18}
\cellbox{lowcell}{20-32}
\cellbox{lowcell}{20-37}
\cellbox{lowcell}{20-42}
\cellbox{lowcell}{21-2}
\cellbox{lowcell}{21-3}
\cellbox{lowcell}{21-7}
\cellbox{lowcell}{21-12}
\cellbox{lowcell}{21-13}
\cellbox{lowcell}{21-17}
\cellbox{lowcell}{21-18}
\cellbox{lowcell}{21-27}
\cellbox{lowcell}{21-32}
\cellbox{lowcell}{21-37}
\cellbox{lowcell}{21-38}
\cellbox{lowcell}{21-42}
\cellbox{lowcell}{22-2}
\cellbox{lowcell}{22-7}
\cellbox{lowcell}{22-22}
\cellbox{lowcell}{22-27}
\cellbox{lowcell}{22-32}
\cellbox{lowcell}{22-37}
\cellbox{lowcell}{22-42}
\cellbox{invrow}{17-4}
\cellbox{invrow}{17-5}
\cellbox{invrow}{17-9}
\cellbox{invrow}{17-10}
\cellbox{invrow}{17-14}
\cellbox{invrow}{17-15}
\cellbox{invrow}{17-19}
\cellbox{invrow}{17-20}
\cellbox{invrow}{17-24}
\cellbox{invrow}{17-25}
\cellbox{invrow}{17-29}
\cellbox{invrow}{17-30}
\cellbox{invrow}{17-34}
\cellbox{invrow}{17-35}
\cellbox{invrow}{17-39}
\cellbox{invrow}{17-40}
\cellbox{invrow}{17-44}
\cellbox{invrow}{17-45}
\cellbox{invrow}{19-4}
\cellbox{invrow}{19-5}
\cellbox{invrow}{19-9}
\cellbox{invrow}{19-10}
\cellbox{invrow}{19-14}
\cellbox{invrow}{19-15}
\cellbox{invrow}{19-19}
\cellbox{invrow}{19-20}
\cellbox{invrow}{19-24}
\cellbox{invrow}{19-25}
\cellbox{invrow}{19-29}
\cellbox{invrow}{19-30}
\cellbox{invrow}{19-34}
\cellbox{invrow}{19-35}
\cellbox{invrow}{19-39}
\cellbox{invrow}{19-40}
\cellbox{invrow}{19-44}
\cellbox{invrow}{19-45}
\cellbox{invrow}{20-4}
\cellbox{invrow}{20-5}
\cellbox{invrow}{20-9}
\cellbox{invrow}{20-10}
\cellbox{invrow}{20-14}
\cellbox{invrow}{20-15}
\cellbox{invrow}{20-19}
\cellbox{invrow}{20-20}
\cellbox{invrow}{20-24}
\cellbox{invrow}{20-25}
\cellbox{invrow}{20-29}
\cellbox{invrow}{20-30}
\cellbox{invrow}{20-34}
\cellbox{invrow}{20-35}
\cellbox{invrow}{20-39}
\cellbox{invrow}{20-40}
\cellbox{invrow}{20-44}
\cellbox{invrow}{20-45}
\cellbox{invrow}{21-4}
\cellbox{invrow}{21-5}
\cellbox{invrow}{21-9}
\cellbox{invrow}{21-10}
\cellbox{invrow}{21-14}
\cellbox{invrow}{21-15}
\cellbox{invrow}{21-19}
\cellbox{invrow}{21-20}
\cellbox{invrow}{21-24}
\cellbox{invrow}{21-25}
\cellbox{invrow}{21-29}
\cellbox{invrow}{21-30}
\cellbox{invrow}{21-34}
\cellbox{invrow}{21-35}
\cellbox{invrow}{21-39}
\cellbox{invrow}{21-40}
\cellbox{invrow}{21-44}
\cellbox{invrow}{21-45}
\cellbox{invrow}{22-4}
\cellbox{invrow}{22-5}
\cellbox{invrow}{22-9}
\cellbox{invrow}{22-10}
\cellbox{invrow}{22-14}
\cellbox{invrow}{22-15}
\cellbox{invrow}{22-19}
\cellbox{invrow}{22-20}
\cellbox{invrow}{22-24}
\cellbox{invrow}{22-25}
\cellbox{invrow}{22-29}
\cellbox{invrow}{22-30}
\cellbox{invrow}{22-34}
\cellbox{invrow}{22-35}
\cellbox{invrow}{22-39}
\cellbox{invrow}{22-40}
\cellbox{invrow}{22-44}
\cellbox{invrow}{22-45}
\cellbox{invrow}{23-4}
\cellbox{invrow}{23-5}
\cellbox{invrow}{23-9}
\cellbox{invrow}{23-10}
\cellbox{invrow}{23-14}
\cellbox{invrow}{23-15}
\cellbox{invrow}{23-19}
\cellbox{invrow}{23-20}
\cellbox{invrow}{23-24}
\cellbox{invrow}{23-25}
\cellbox{invrow}{23-29}
\cellbox{invrow}{23-30}
\cellbox{invrow}{23-34}
\cellbox{invrow}{23-35}
\cellbox{invrow}{23-39}
\cellbox{invrow}{23-40}
\cellbox{invrow}{23-44}
\cellbox{invrow}{23-45}
\end{NiceTabular}
\end{sidewaystable*}

%% file: appA_scaling_1.tex
\begin{table*}[p]
\centering
\caption{Accuracy drop (\gap{} $\downarrow$) by number of options, CLINC150, in points. $n{=}3$ training seeds; $\pm$: standard deviation, descriptive only.}
\label{tab:app-scaling-1}
\scriptsize
\setlength{\tabcolsep}{2.5pt}
\begin{tabular}{lccccccc}
\toprule
 & & \multicolumn{6}{c}{Number of options $K$} \\
\cmidrule(lr){3-8}
Method & Passes / q & 4 & 10 & 20 & 50 & 100 & 151 \\
\midrule
\multicolumn{8}{l}{\textit{Baseline}} \\
Baseline model 1.7B & 1 & 0.0$\pm$0.0 & 0.4$\pm$0.1 & 1.2$\pm$0.6 & 2.3$\pm$0.9 & 8.5$\pm$2.4 & 10.7$\pm$1.2 \\
\midrule
\multicolumn{8}{l}{\textit{Post-hoc calibration (one pass; applied to the baseline model)}} \\
Prior correction (PriDe, 5\%)$^{a}$ & $1{+}\frac{K-1}{20}$ & 0.0$\pm$0.0 & 0.4$\pm$0.1 & 1.4$\pm$0.4 & 3.8$\pm$1.3 & 11.9$\pm$2.2 & 16.9$\pm$3.2 \\
Per-position temperature scaling$^{a}$ & 1 & 0.0$\pm$0.0 & 0.4$\pm$0.1 & 1.2$\pm$0.6 & 2.1$\pm$0.6 & 8.3$\pm$2.6 & 10.7$\pm$1.2 \\
\midrule
\multicolumn{8}{l}{\textit{Training-time augmentation (same base and data as the baseline model)}} \\
Augmentation (2 ep.) & 1 & 0.0$\pm$0.0 & 0.8$\pm$0.4 & 1.5$\pm$0.9 & 3.1$\pm$1.5 & 8.0$\pm$1.5 & 10.8$\pm$2.1 \\
Augmentation (4 ep.) & 1 & 0.3$\pm$0.2 & 0.8$\pm$0.5 & 2.8$\pm$0.7 & 2.7$\pm$1.1 & 7.0$\pm$0.8 & 9.6$\pm$1.5 \\
Augmentation (4 ep.; fixed) & 1 & 0.1$\pm$0.1 & 0.8$\pm$0.4 & 2.6$\pm$0.3 & 5.6$\pm$0.5 & 12.2$\pm$1.1 & 16.1$\pm$1.8 \\
\midrule
\multicolumn{8}{l}{\textit{Test-time ensembling (applied to the baseline model)}} \\
Random-permutation ensemble (avg, 2)$^{b}$ & 2 & 0.0$\pm$0.0 & 0.3$\pm$0.2 & 0.8$\pm$0.1 & 1.0$\pm$0.4 & 4.6$\pm$1.4 & 6.7$\pm$0.8 \\
Random-permutation ensemble (avg, 4)$^{b}$ & 4 & 0.0$\pm$0.0 & 0.2$\pm$0.2 & 0.5$\pm$0.1 & 0.6$\pm$0.6 & 3.0$\pm$1.0 & 4.7$\pm$0.7 \\
Random-permutation ensemble (avg, 8)$^{b}$ & 8 & 0.0$\pm$0.0 & 0.1$\pm$0.1 & 0.4$\pm$0.1 & 0.5$\pm$0.7 & 2.2$\pm$0.4 & 3.3$\pm$0.3 \\
Random-permutation ensemble (vote, 2)$^{b}$ & 2 & 0.0$\pm$0.0 & 0.3$\pm$0.2 & 0.8$\pm$0.1 & 1.0$\pm$0.4 & 4.6$\pm$1.4 & 6.8$\pm$0.8 \\
Random-permutation ensemble (vote, 4)$^{b}$ & 4 & 0.0$\pm$0.0 & 0.2$\pm$0.2 & 0.6$\pm$0.1 & 0.7$\pm$0.6 & 3.2$\pm$1.2 & 5.0$\pm$0.8 \\
Random-permutation ensemble (vote, 8)$^{b}$ & 8 & 0.0$\pm$0.0 & 0.1$\pm$0.1 & 0.4$\pm$0.1 & 0.5$\pm$0.7 & 2.5$\pm$0.5 & 3.7$\pm$0.2 \\
Cyclic-permutation ensemble (avg)$^{c}$ & $K$ & 0.0$\pm$0.0 & 0.2$\pm$0.3 & 0.2$\pm$0.1 & 0.6$\pm$0.7 & \textit{n/a} & \textit{n/a} \\
Cyclic-permutation ensemble (L0)$^{c}$ & $K$ & 0.0$\pm$0.0 & 0.1$\pm$0.2 & 0.8$\pm$0.2 & 5.1$\pm$1.9 & \textit{n/a} & \textit{n/a} \\
Cyclic-permutation ensemble (vote)$^{c}$ & $K$ & 0.0$\pm$0.0 & 0.0$\pm$0.0 & 0.3$\pm$0.1 & 0.6$\pm$0.6 & \textit{n/a} & \textit{n/a} \\
Cyclic-permutation ensemble (L0, vote)$^{c}$ & $K$ & 0.0$\pm$0.0 & 0.1$\pm$0.2 & 0.8$\pm$0.2 & 5.4$\pm$1.7 & \textit{n/a} & \textit{n/a} \\
Filtered cyclic-permutation ensemble$^{f}$ & $1{+}|C|$ & 0.0 & 0.0 & 0.0 & 0.0 & 0.0 & 0.0 \\
\midrule
\multicolumn{8}{l}{\textit{Architectural designs}} \\
Separate option encoding & $K{+}1$ & 0.0 & 0.0 & 0.0 & 0.0 & 0.0 & 0.0 \\
Separate option encoding (our prompt, Jeva's scoring head) & $K{+}1$ & 0.0 & 0.0 & 0.0 & 0.0 & 0.0 & 0.0 \\
Separate option encoding (Jeva's implementation) & $K{+}1$ & 0.0 & 0.0 & 0.0 & 0.0 & 0.0 & 0.0 \\
\Method{} (test-time) & 1 & 0.0$\pm$0.0 & 0.0$\pm$0.0 & 0.0$\pm$0.0 & 0.0$\pm$0.0 & 0.0$\pm$0.0 & 0.0$\pm$0.0 \\
\Method{} 1.7B & 1 & 0.0 & 0.0 & 0.0 & 0.0 & 0.0 & 0.0 \\
\Method{} 4B$^\dagger$ & 1 & 0.0 & 0.0 & 0.0 & 0.0 & 0.0 & 0.0 \\
\midrule
\multicolumn{8}{l}{\textit{Open-weight models (released checkpoints; different bases and data)}} \\
Tacit 1.7B$^\dagger$ & 1 & 4.5 & 10.2 & 14.8 & \textit{n/a} & \textit{n/a} & \textit{n/a} \\
d3-nano 2.2B$^\dagger$ & 1 & 0.6 & 0.2 & 1.5 & 2.3 & 7.5 & 11.1 \\
Intern-Decision 4B$^\dagger$ & 1 & 2.8 & 7.0 & 11.1 & 23.6 & \textit{n/a} & \textit{n/a} \\
Jeva 4B$^\dagger$ & $K{+}1$ & 0.0 & 0.0 & 0.0 & 0.0 & 0.0 & 0.0 \\
Kev 4B$^\dagger$ & 1 & 2.9 & 6.3 & 12.2 & 26.0 & 32.8 & 37.8 \\
Tacit 4B$^\dagger$ & 1 & 0.0 & 5.3 & 14.5 & \textit{n/a} & \textit{n/a} & \textit{n/a} \\
Elephant-1 9B$^\dagger$ & 1 & 0.0 & 0.4 & 0.4 & 0.9 & 3.9 & 6.7 \\
Mammoth-1 27B$^\dagger$ & 1 & 0.0 & 0.0 & 0.0 & 0.0 & 1.5 & 2.6 \\
\midrule
\multicolumn{8}{l}{\textit{Hosted APIs (each on the items it answered)}} \\
Jev API & 1 & 0.5 & 0.9 & 2.2 & 7.4 & 11.6 & 17.0 \\
OpenAI API & 1 & 2.1 & 2.1 & 4.0 & 19.1 & 38.2 & 47.8 \\
\bottomrule
\end{tabular}
\end{table*}

%% file: appA_scaling_2.tex
\begin{table*}[p]
\centering
\caption{Accuracy drop (\gap{} $\downarrow$) by number of options, MASSIVE, in points. $n{=}3$ training seeds; $\pm$: standard deviation, descriptive only.}
\label{tab:app-scaling-2}
\scriptsize
\setlength{\tabcolsep}{2.5pt}
\begin{tabular}{lcccccc}
\toprule
 & & \multicolumn{5}{c}{Number of options $K$} \\
\cmidrule(lr){3-7}
Method & Passes / q & 4 & 10 & 20 & 40 & 60 \\
\midrule
\multicolumn{7}{l}{\textit{Baseline}} \\
Baseline model 1.7B & 1 & 0.8$\pm$0.5 & 1.9$\pm$0.6 & 4.8$\pm$1.0 & 8.2$\pm$1.1 & 11.0$\pm$2.7 \\
\midrule
\multicolumn{7}{l}{\textit{Post-hoc calibration (one pass; applied to the baseline model)}} \\
Prior correction (PriDe, 5\%)$^{a}$ & $1{+}\frac{K-1}{20}$ & 0.7$\pm$0.4 & 1.6$\pm$0.4 & 5.1$\pm$1.0 & 10.2$\pm$1.8 & 15.2$\pm$3.1 \\
Per-position temperature scaling$^{a}$ & 1 & 0.5$\pm$0.3 & 1.5$\pm$0.5 & 4.7$\pm$1.5 & 7.7$\pm$1.1 & 11.2$\pm$2.8 \\
\midrule
\multicolumn{7}{l}{\textit{Training-time augmentation (same base and data as the baseline model)}} \\
Augmentation (2 ep.) & 1 & 0.6$\pm$0.6 & 2.5$\pm$0.2 & 6.2$\pm$0.9 & 8.6$\pm$1.0 & 11.8$\pm$1.7 \\
Augmentation (4 ep.) & 1 & 0.9$\pm$0.8 & 2.5$\pm$0.7 & 9.1$\pm$2.0 & 9.4$\pm$0.3 & 15.1$\pm$1.8 \\
Augmentation (4 ep.; fixed) & 1 & 1.1$\pm$0.1 & 5.4$\pm$0.7 & 10.8$\pm$1.1 & 16.4$\pm$1.2 & 19.5$\pm$1.7 \\
\midrule
\multicolumn{7}{l}{\textit{Test-time ensembling (applied to the baseline model)}} \\
Random-permutation ensemble (avg, 2)$^{b}$ & 2 & 0.5$\pm$0.2 & 1.3$\pm$0.3 & 2.6$\pm$0.5 & 4.8$\pm$0.9 & 6.8$\pm$1.5 \\
Random-permutation ensemble (avg, 4)$^{b}$ & 4 & 0.4$\pm$0.3 & 0.9$\pm$0.2 & 1.9$\pm$0.7 & 3.3$\pm$0.4 & 4.1$\pm$0.8 \\
Random-permutation ensemble (avg, 8)$^{b}$ & 8 & 0.1$\pm$0.1 & 0.7$\pm$0.3 & 1.3$\pm$0.6 & 2.1$\pm$0.9 & 2.8$\pm$1.0 \\
Random-permutation ensemble (vote, 2)$^{b}$ & 2 & 0.5$\pm$0.2 & 1.3$\pm$0.3 & 2.6$\pm$0.5 & 4.8$\pm$0.9 & 6.8$\pm$1.5 \\
Random-permutation ensemble (vote, 4)$^{b}$ & 4 & 0.4$\pm$0.3 & 0.9$\pm$0.2 & 2.0$\pm$0.6 & 3.3$\pm$0.4 & 4.3$\pm$1.0 \\
Random-permutation ensemble (vote, 8)$^{b}$ & 8 & 0.2$\pm$0.1 & 0.7$\pm$0.3 & 1.4$\pm$0.5 & 2.4$\pm$0.8 & 3.1$\pm$1.1 \\
Cyclic-permutation ensemble (avg)$^{c}$ & $K$ & 0.1$\pm$0.2 & 0.9$\pm$0.5 & 2.2$\pm$0.8 & 3.2$\pm$1.6 & 3.9$\pm$1.2 \\
Cyclic-permutation ensemble (L0)$^{c}$ & $K$ & 0.5$\pm$0.3 & 0.7$\pm$0.3 & 3.1$\pm$0.3 & 7.0$\pm$1.7 & 11.7$\pm$1.9 \\
Cyclic-permutation ensemble (vote)$^{c}$ & $K$ & 0.3$\pm$0.2 & 1.1$\pm$0.6 & 1.9$\pm$0.5 & 3.9$\pm$1.1 & 4.1$\pm$1.6 \\
Cyclic-permutation ensemble (L0, vote)$^{c}$ & $K$ & 0.5$\pm$0.2 & 0.9$\pm$0.5 & 3.1$\pm$0.6 & 7.2$\pm$2.3 & 13.0$\pm$2.1 \\
Filtered cyclic-permutation ensemble$^{f}$ & $1{+}|C|$ & 0.0 & 0.0 & 0.0 & 0.0 & 0.0 \\
\midrule
\multicolumn{7}{l}{\textit{Architectural designs}} \\
Separate option encoding & $K{+}1$ & 0.0 & 0.0 & 0.0 & 0.0 & 0.0 \\
Separate option encoding (our prompt, Jeva's scoring head) & $K{+}1$ & 0.0 & 0.0 & 0.0 & 0.0 & 0.0 \\
Separate option encoding (Jeva's implementation) & $K{+}1$ & 0.0 & 0.0 & 0.0 & 0.0 & 0.0 \\
\Method{} (test-time) & 1 & 0.0$\pm$0.0 & 0.0$\pm$0.0 & 0.0$\pm$0.0 & 0.0$\pm$0.0 & 0.0$\pm$0.0 \\
\Method{} 1.7B & 1 & 0.0 & 0.0 & 0.0 & 0.0 & 0.0 \\
\Method{} 4B$^\dagger$ & 1 & 0.0 & 0.0 & 0.0 & 0.0 & 0.0 \\
\midrule
\multicolumn{7}{l}{\textit{Open-weight models (released checkpoints; different bases and data)}} \\
Tacit 1.7B$^\dagger$ & 1 & 3.6 & 13.5 & 22.7 & \textit{n/a} & \textit{n/a} \\
d3-nano 2.2B$^\dagger$ & 1 & 0.8 & 0.8 & 2.0 & 6.3 & 9.5 \\
Intern-Decision 4B$^\dagger$ & 1 & 3.1 & 6.8 & 14.8 & 26.4 & 68.2 \\
Jeva 4B$^\dagger$ & $K{+}1$ & 0.0 & 0.0 & 0.0 & 0.0 & 0.0 \\
Kev 4B$^\dagger$ & 1 & 2.9 & 8.7 & 11.8 & 21.5 & 27.7 \\
Tacit 4B$^\dagger$ & 1 & 3.0 & 10.5 & 20.7 & \textit{n/a} & \textit{n/a} \\
Elephant-1 9B$^\dagger$ & 1 & 0.5 & 0.6 & 1.6 & 2.9 & 3.6 \\
Mammoth-1 27B$^\dagger$ & 1 & 0.0 & 0.0 & 0.9 & 2.8 & 2.9 \\
\midrule
\multicolumn{7}{l}{\textit{Hosted APIs (each on the items it answered)}} \\
Jev API & 1 & 0.9 & 2.7 & 6.2 & 9.6 & 12.3 \\
OpenAI API & 1 & 1.9 & 6.5 & 10.6 & 28.6 & 33.5 \\
\bottomrule
\end{tabular}
\end{table*}

%% file: appA_m2w_tasks.tex
\begin{table*}[p]
\centering
\caption{Mind2Web task success in \%: original order, average accuracy (\avgacc), worst-case accuracy (\worstacc), order variance (Var.) and accuracy drop (\gap). $n{=}3$ training seeds; $\pm$: standard deviation, descriptive only. Var.: mean over seeds; standard deviation at most 0.3.}
\label{tab:app-m2w-tasks}
\small
\setlength{\tabcolsep}{2.5pt}
\begin{tabular}{lcccccc}
\toprule
Method & Passes / q & Original & \avgacc & \worstacc & Var. & \gap \\
\midrule
\multicolumn{7}{l}{\textit{Baseline}} \\
Baseline model 1.7B & 1 & 2.3$\pm$0.7 & 2.8$\pm$0.4 & 1.3$\pm$0.3 & 0.5 & 1.5$\pm$0.5 \\
\midrule
\multicolumn{7}{l}{\textit{Post-hoc calibration (one pass; applied to the baseline model)}} \\
Prior correction (PriDe, 5\%)$^{a}$ & $1{+}\frac{K-1}{20}$ & 2.2$\pm$0.7 & 2.8$\pm$0.4 & 1.5$\pm$0.4 & 0.5 & 1.3$\pm$0.4 \\
Per-position temperature scaling$^{a}$ & 1 & 2.1$\pm$0.5 & 2.8$\pm$0.4 & 1.3$\pm$0.3 & 0.5 & 1.4$\pm$0.5 \\
\midrule
\multicolumn{7}{l}{\textit{Training-time augmentation (same base and data as the baseline model)}} \\
Augmentation (2 ep.) & 1 & 2.2$\pm$0.8 & 2.9$\pm$0.3 & 1.8$\pm$0.4 & 0.4 & 1.1$\pm$0.1 \\
Augmentation (4 ep.) & 1 & 2.6$\pm$0.2 & 2.7$\pm$0.3 & 1.7$\pm$0.0 & 0.5 & 1.0$\pm$0.3 \\
Augmentation (4 ep.; fixed) & 1 & 2.4$\pm$0.5 & 3.3$\pm$0.1 & 1.6$\pm$0.4 & 0.8 & 1.7$\pm$0.5 \\
\midrule
\multicolumn{7}{l}{\textit{Test-time ensembling (applied to the baseline model)}} \\
Random-permutation ensemble (avg, 2)$^{b}$ & 2 & -- & 2.8$\pm$0.2 & 1.7$\pm$0.2 & 0.4 & 1.1$\pm$0.3 \\
Random-permutation ensemble (avg, 4)$^{b}$ & 4 & -- & 2.9$\pm$0.2 & 2.0$\pm$0.3 & 0.3 & 0.9$\pm$0.3 \\
Random-permutation ensemble (avg, 8)$^{b}$ & 8 & -- & 3.0$\pm$0.3 & 2.3$\pm$0.4 & 0.2 & 0.7$\pm$0.4 \\
Random-permutation ensemble (vote, 2)$^{b}$ & 2 & -- & 2.8$\pm$0.2 & 1.7$\pm$0.2 & 0.4 & 1.1$\pm$0.3 \\
Random-permutation ensemble (vote, 4)$^{b}$ & 4 & -- & 2.9$\pm$0.2 & 1.9$\pm$0.3 & 0.3 & 1.0$\pm$0.3 \\
Random-permutation ensemble (vote, 8)$^{b}$ & 8 & -- & 3.0$\pm$0.3 & 2.2$\pm$0.3 & 0.3 & 0.8$\pm$0.3 \\
Cyclic-permutation ensemble (avg)$^{c}$ & $K$ & -- & 2.9$\pm$0.2 & 2.0$\pm$0.3 & 0.3 & 0.9$\pm$0.4 \\
Cyclic-permutation ensemble (L0)$^{c}$ & $K$ & -- & 3.0$\pm$0.1 & 2.2$\pm$0.5 & 0.3 & 0.7$\pm$0.6 \\
Cyclic-permutation ensemble (vote)$^{c}$ & $K$ & -- & 3.0$\pm$0.2 & 2.0$\pm$0.3 & 0.3 & 1.0$\pm$0.5 \\
Cyclic-permutation ensemble (L0, vote)$^{c}$ & $K$ & -- & 3.0$\pm$0.2 & 1.9$\pm$0.4 & 0.3 & 1.1$\pm$0.5 \\
Filtered cyclic-permutation ensemble$^{f}$ & $1{+}|C|$ & 3.3$\pm$0.7 & 3.3$\pm$0.7 & 3.3$\pm$0.7 & 0.0 & 0.0 \\
\midrule
\multicolumn{7}{l}{\textit{Architectural designs}} \\
Separate option encoding & $K{+}1$ & 2.6$\pm$0.5 & 2.6$\pm$0.5 & 2.6$\pm$0.5 & 0.0 & 0.0 \\
Separate option encoding (our prompt, Jeva's scoring head) & $K{+}1$ & 1.2$\pm$0.7 & 1.2$\pm$0.7 & 1.2$\pm$0.7 & 0.0 & 0.0 \\
Separate option encoding (Jeva's implementation) & $K{+}1$ & 1.6$\pm$0.7 & 1.6$\pm$0.7 & 1.6$\pm$0.7 & 0.0 & 0.0 \\
\Method{} (test-time) & 1 & 2.3$\pm$0.6 & 2.3$\pm$0.6 & 2.2$\pm$0.5 & 0.0 & 0.1$\pm$0.2 \\
\Method{} 1.7B & 1 & 2.8$\pm$0.2 & 2.8$\pm$0.2 & 2.8$\pm$0.2 & 0.0 & 0.0 \\
\Method{} 4B$^\dagger$ & 1 & 5.0 & 5.0 & 5.0 & 0.0 & 0.0 \\
\midrule
\multicolumn{7}{l}{\textit{Open-weight models (released checkpoints; different bases and data)}} \\
Tacit 1.7B$^\dagger$ & 1 & 1.7 & 1.5 & 0.3 & 0.5 & 1.2 \\
d3-nano 2.2B$^\dagger$ & 1 & 4.3 & 4.5 & 2.0 & 1.3 & 2.5 \\
Intern-Decision 4B$^\dagger$ & 1 & 5.0 & 4.8 & 1.0 & 1.7 & 3.8 \\
Jeva 4B$^\dagger$ & $K{+}1$ & 1.7 & 1.7 & 1.7 & 0.0 & 0.0 \\
Kev 4B$^\dagger$ & 1 & 4.7 & 4.9 & 2.0 & 1.1 & 2.9 \\
Tacit 4B$^\dagger$ & 1 & 3.0 & 3.0 & 1.3 & 0.7 & 1.6 \\
d3-mini 4.5B$^\dagger$ & 1 & 7.7 & 7.9 & 4.0 & 1.4 & 3.9 \\
Elephant-1 9B$^\dagger$ & 1 & 10.3 & 10.7 & 8.3 & 0.8 & 2.3 \\
Mammoth-1 27B$^\dagger$ & 1 & 14.3 & 14.2 & 10.3 & 1.2 & 3.9 \\
\midrule
\multicolumn{7}{l}{\textit{Hosted APIs (each on the items it answered)}} \\
Jev API & 1 & 9.7 & 9.8 & 6.0 & 1.2 & 3.8 \\
OpenAI API & 1 & 8.3 & 8.9 & 3.0 & 2.3 & 5.9 \\
\bottomrule
\end{tabular}
\end{table*}

%% file: appA_all_orders.tex
\begin{table}[t]
\centering
\caption{How tight the Latin-square estimate of worst-case accuracy (\worstacc) is: \worstacc{} in \% over the $K$ Latin-square orders (Latin) and over all or many orders of the same items, and their difference in points.}
\label{tab:app-all-orders}
\scriptsize
\setlength{\tabcolsep}{2.5pt}
\begin{tabular}{lccccccccc}
\toprule
 & \multicolumn{3}{c}{MMLU ($K{=}4$, 300 items)} & \multicolumn{3}{c}{ARC ($K{=}4$, 300 items)} & \multicolumn{3}{c}{MMLU-Pro ($K{=}10$, 100 items)} \\
\cmidrule(lr){2-4}\cmidrule(lr){5-7}\cmidrule(lr){8-10}
System & Latin & all 24 & Gap & Latin & all 24 & Gap & Latin & 50 random & Gap \\
\midrule
Kev 4B$^\dagger$ & 61.7 & 56.3 & 5.3 & 87.0 & 84.3 & 2.7 & 32.0 & 30.0 & 2.0 \\
Kev 9B$^\dagger$ & 65.3 & 61.0 & 4.3 & 91.0 & 88.7 & 2.3 & 39.0 & 35.0 & 4.0 \\
\bottomrule
\end{tabular}
\end{table}

%% file: B_method_details.tex
\section{Method details}
\label{app:methods}

\para{Random- and cyclic-permutation ensembles.}
The random-permutation ensemble scores the question under $m$ random option orders and averages the $m$ distributions over the options.
The cyclic-permutation ensemble does the same over the $K$ cyclic shifts of the given order, so every option appears once at every position.
With the L0 reducer of AnyJev \citep{anyjev2026}, each shift's distribution over positions is first divided by that shift's position prior, the mean distribution over the questions of the set raised to the power $0.75$, and the corrected distributions are combined by their geometric mean.

\para{PriDe.}
Let $p^{(r)}_n\in\Delta^{K-1}$ be the baseline model's distribution over positions for question $n$ in the $r$-th of its $K$ Latin-square orders.
On a random 5\% of the questions, $S$, each question's prior is the normalized geometric mean of its $K$ distributions, and the set's prior is their average:
\begin{equation*}
  \hat\pi_n \propto \exp\Big(\tfrac1K\textstyle\sum_{r=1}^{K}\log p^{(r)}_n\Big),\qquad \hat\pi=\tfrac{1}{|S|}\textstyle\sum_{n\in S}\hat\pi_n .
\end{equation*}
On every other question, the corrected distribution is $\tilde p_k\propto p_k/\hat\pi_k$.
Metrics are averaged over 5 random draws of $S$; the cost is one pass per question, plus $K-1$ more on each question of $S$.

\para{Per-position temperature scaling.}
The corrected distribution is $\tilde p_k=\mathrm{softmax}_k(\log p_k/\tau_k)$.
The temperatures of each set (and each $K$) are fitted with L-BFGS on all Latin-square orders of the questions of the other fold, minimizing
\begin{equation*}
  -\tfrac1N\textstyle\sum_{n}\log\tilde p_{y_n} + \lambda\,\tfrac1K\textstyle\sum_{k}(\log\tau_k)^2,\qquad \lambda=0.01,
\end{equation*}
where $y_n$ is the position of the gold option in answer $n$.
The questions are split in two halves; the temperatures fitted on one half are applied to the other, and the halves are swapped.

\para{Filtered cyclic-permutation ensemble.}
At the filter stage, the model scores each option alone in the yes/no question format.
The score is the log-probability of yes minus that of no, and the scores are normalized with a softmax.
The filter keeps the set $C$ of top options whose combined mass is over a threshold $\alpha$, with $|C|\ge k_{\min}$.
The options in $C$ are then ranked by the cyclic-permutation ensemble over the shifts of $C$ alone (geometric mean, no position prior) and share the filter's mass on $C$ in proportion to that ranking; options outside $C$ keep their filter probability.
As in Runoff \citep{huang2026runoff}, we use $\alpha=0.99$ and $k_{\min}=3$, and both stages use the baseline model.

\para{Tree attention at test time.}
To separate the effect of the mask from that of training under it, we also apply tree attention to the trained baseline at inference, without retraining, as Set-Based Prompting does for pretrained language models \citep{mcilroy2024order}.
It is listed as ``\Method{} (test-time)'' in the tables.

%% file: C_datasets.tex
\section{Datasets and option-count sweeps}
\label{app:datasets}

\para{Evaluation sets.}
\emph{Multiple choice:} ARC-Challenge \citep{clark2018arc} (1,165 test questions, $K=4$), MMLU \citep{hendrycks2021mmlu} (1,909 questions, $K=4$) and MMLU-Pro \citep{wang2024mmlupro} (1,659 questions, up to $K=10$).
\emph{Intent classification:} every intent label of the dataset is an option. BANKING77 \citep{casanueva2020banking} has 300 test utterances with $K=77$, CLINC150 \citep{larson2019clinc} 200 test utterances with $K=151$ (150 intents and out-of-scope), and MASSIVE \citep{fitzgerald2023massive} 200 English test utterances with $K=60$; MASSIVE utterances whose text also appears in training are excluded.
\emph{Agentic:} Mind2Web \citep{deng2023mind2web}, 2,118 steps from 300 test tasks (100 from each of its cross-task, cross-website and cross-domain splits).
At each step the context holds the task, the pruned web page and the gold previous actions, as in the standard Mind2Web step evaluation, and the $K=10$ options are the gold element and the nine highest-ranked other elements under the candidate ranker released with the dataset.
BANKING77, CLINC150, ARC-Challenge and Mind2Web contribute their training splits to the training corpus (Appendix~\ref{app:training}), so these four sets are in-domain; MMLU, MMLU-Pro and MASSIVE are out-of-domain.

\para{Option-count sweeps.}
To vary $K$ with the questions fixed, we use the same 200 CLINC150 questions at $K=4,10,20,50,100,151$, and the same 200 MASSIVE utterances at $K=4,10,20,40,60$.
The option sets are nested: each smaller set contains the gold option and is a subset of every larger one, so a change across $K$ comes from the added options alone.

\para{Evaluation orders.}
Each question is evaluated in the $K$ orders of a random Latin square: order $s$ lists option $a_{b((c_j+s) \bmod K)}$ at position $j$, for random permutations $b$ and $c$ of the $K$ indices drawn from a seed fixed per question.
Every option thus appears exactly once at every position, and all arms see the same orders.
To check how close the worst case over these $K$ orders comes to the worst case over all orders, 300 MMLU and 300 ARC-Challenge questions are also scored in all 24 orders and 100 MMLU-Pro questions in 50 random orders (\cref{tab:app-all-orders}).

%% file: D_setup.tex
\section{Experimental setup in detail}
\label{app:setup}

\para{Models.}
The baseline model is Qwen3-1.7B-Base fine-tuned for one epoch on the corpus of Appendix~\ref{app:training}, with the read-out head of \cref{eq:readout}.
It reads the context and the options as one causal sequence with sequential position ids, so each option attends to every option listed before it.
The input is wrapped in special tokens: the context in \texttt{<context>}\ldots\texttt{</context>}, the question text after \texttt{<question>}, each option in \texttt{<choice>}\ldots\texttt{</choice>}, and a final \texttt{</question>} token after the options.
The read-out head takes the hidden state of \texttt{</question>}, which attends to the context and all options, as the query $h^{\pi}_{\mathrm{r}}$, and the hidden state of each option's \texttt{</choice>} as its key $h^{\pi}_{a}$ (\cref{eq:readout}); the special tokens have their own learned embeddings.
Every trained arm uses the same corpus, schedule and optimizer, and differs from the baseline in one respect: the option orders and number of epochs in training (augmentation), the attention mask and position ids (\method{}), or how the inputs are encoded (separate encoding).
Augmentation draws a fresh option order for every example in every epoch and trains for 2 or 4 epochs; to separate the effect of the orders from that of longer training, a control trains 4 epochs on one fixed order.
\Method{} follows the layout of Set-Based Prompting \citep{mcilroy2024order}: every option starts right after the context, and the read-out token, placed after the longest option, attends to all options.
Separate encoding encodes the context and each option in their own passes with the baseline's prompt format and cosine head; its ablation replaces the head with Jeva's decision head \citep{li2026jeva}, and a control trains Jeva's own implementation on our corpus (appendix tables only).
For the comparison with released models (\cref{sec:sanity-check}), \method{} 4B is trained from Qwen3-4B-Base \citep{yang2025qwen3} with the same recipe and data.

\para{Settings of each variant.}
PriDe estimates its prior from 5\% of the questions, and per-position temperature scaling is fitted with two-fold cross-fitting (Appendix~\ref{app:methods}).
The random-permutation ensemble uses $m=2,4,8$ orders; the cyclic-permutation ensemble with L0 uses AnyJev's exponent of 0.75 \citep{anyjev2026}; the filtered ensemble uses $\alpha=0.99$ and $k_{\min}=3$, as in Runoff \citep{huang2026runoff}.

\para{Evaluation protocol.}
Every method is evaluated on the $K$ Latin-square orders of each question (Appendix~\ref{app:datasets}), which are shared by all arms, and its cost is the number of forward passes per evaluation.
The cyclic-permutation ensemble combines the $K$ cyclic shifts of each Latin-square order, $K^2$ passes per question, so we run it only up to $K=60$.
The random-permutation ensemble draws new random orders for every evaluation, and its metrics are averaged over five draws.
Released models are evaluated with their own code (Kev through its released server), and hosted APIs through their public endpoints (Appendix~\ref{app:api}).

\para{Hardware.}
All models run in bf16 on NVIDIA B300, B200 or RTX A6000 GPUs (Appendix~\ref{app:precision}).

%% file: E_training.tex
\section{Training details}
\label{app:training}

\para{Training corpus.}
Every trained arm uses one corpus of 891,107 training examples (108M tokens) from 48 public datasets in eight domains (\cref{tab:corpus}): knowledge and reasoning QA, commonsense inference, reading comprehension, inference over rules, text and intent classification, agents, tools and dialogue, response judging and safety, and synthetic procedural decisions.
Each source is converted from its training split into one form: a context, one or more questions about it, and each question's options with one gold option. An example is one question: the model answers it on its own from the context and gets one loss for it; a context with several questions (e.g. a RACE passage) gives several examples.
Options per question range from 2 to 255 (median 3); classification and agent sources give many-option questions (BANKING77 77, CLINC150 151, RICO-SCA up to 223).
We then removed training questions that overlap an evaluation benchmark (same question, same context, or a shared 13-word $n$-gram, ignoring templates and boilerplate), dropped duplicate questions within each source, and capped each source at a fixed number of questions, which also excluded eight further sources; 891,107 of 1,618,199 examples remain.
Finally, the options of every question were shuffled once with a seeded permutation, so the gold option is uniformly distributed over positions.
A separate check of the final corpus against every evaluation set found no shared question or 13-word $n$-gram, apart from generic wording (three MMLU questions such as ``Which of the following statements is true?'' and one CLINC150 utterance, ``that's not correct'').
BANKING77, CLINC150, ARC-Challenge and Mind2Web contribute their training splits, so these evaluation sets are in-domain; MMLU, MMLU-Pro and MASSIVE do not, although the corpus has related tasks (science and medical multiple choice; voice-assistant intents in CLINC150).
The baseline reads the corpus once in 2,054 steps (200 steps training only the head, then 1,854 of full fine-tuning), each step 8 packed rows of up to 8,192 tokens.
Contexts longer than 6,144 tokens are truncated; this affects 619 contexts, 520 of Mind2Web's 1,394 and 90 of QuALITY's 145.

\input{appE_corpus.tex}

\para{Epochs and seeds.}
Augmentation trains for 2 or 4 epochs, i.e.\ 2 or 4 times the baseline's training steps, and its fixed-order control for 4 epochs; all other trained arms train for one epoch, like the baseline.
All variants are trained in bf16 on one NVIDIA B200 GPU with FlashAttention-4 kernel \citep{zadouri2026flashattention4}.
Every trained 1.7B arm has three training seeds, which set the head initialization, the order of the training data and the per-epoch option reshuffle; the training arguments are otherwise identical.

\para{Training cost.}
We compare training cost by the time per training step with FlashAttention-4, measured on one NVIDIA B200 GPU (PyTorch 2.10, CUDA 13.0, Transformers 4.57, bf16) for training seeds 1 and 2 of the 1.7B arms.
The baseline and \method{} train on identical batches, so their per-step times compare the same work: a step of 434 examples takes about 2.04\,s for the baseline, 2.44\,s for \method{} and 6.40\,s for separate encoding.
\Method{} is slower because its attention mask is executed as several FlashAttention-4 calls per layer whose partial results are merged, instead of one causal call.
Augmentation has the baseline's per-step cost but trains for 2 or 4 epochs, i.e.\ 2 or 4 times as many steps.

%% file: appE_corpus.tex
\begin{table*}[!t]
\centering
\caption{Training corpus: every trained arm uses this one copy (one pass, Appendix~\ref{app:training}). 48 sources in 8 domains, 891{,}107 examples over 757{,}256 distinct contexts, 108M tokens. Example: one question with its options and gold option, one loss each; a context can carry several examples (e.g. RACE passages, CFPB complaints), each answered on its own; options: per example (median, max); tokens: Qwen3 tokens of the context, questions and options, without formatting. Every source contributes its training split only. $^\star$ same dataset as an evaluation set, which is therefore in-domain (BANKING77, CLINC150, ARC-Challenge, Mind2Web); MMLU, MMLU-Pro and MASSIVE have no source here.}
\label{tab:corpus}
\scriptsize
\setlength{\tabcolsep}{3pt}
\begin{tabular}{@{}llrrrr@{}}
\toprule
Source (Hugging Face) & Task & Examples & $K$ (med) & $K$ (max) & Tokens (M) \\
\midrule
\multicolumn{2}{@{}l}{\textit{Knowledge and reasoning QA}} & \textit{204{,}267} & & & \textit{8.8} \\
\quad allenai/sciq & science question & 6{,}000 & 4 & 4 & 0.7 \\
\quad allenai/ai2\_arc (Easy) & grade-school science & 2{,}231 & 4 & 5 & 0.1 \\
\quad allenai/ai2\_arc (Challenge)$^\star$ & grade-school science & 1{,}100 & 4 & 5 & 0.1 \\
\quad allenai/openbookqa & elementary science & 4{,}946 & 4 & 4 & 0.1 \\
\quad tau/commonsense\_qa & commonsense question & 9{,}000 & 5 & 5 & 0.2 \\
\quad openlifescienceai/medmcqa & medical exam question & 150{,}990 & 4 & 4 & 5.8 \\
\quad deepmind/aqua\_rat & math word problem & 30{,}000 & 5 & 5 & 1.9 \\
\multicolumn{2}{@{}l}{\textit{Commonsense inference}} & \textit{91{,}376} & & & \textit{5.3} \\
\quad Rowan/hellaswag & plausible continuation & 12{,}000 & 4 & 4 & 2.2 \\
\quad allenai/winogrande (xl) & fill the blank & 40{,}390 & 2 & 2 & 1.3 \\
\quad ybisk/piqa & physical goal, solution & 15{,}995 & 2 & 2 & 1.0 \\
\quad allenai/social\_i\_qa & social situation & 22{,}991 & 3 & 3 & 0.8 \\
\multicolumn{2}{@{}l}{\textit{Reading comprehension}} & \textit{44{,}445} & & & \textit{6.8} \\
\quad ehovy/race & exam passage question & 20{,}000 & 4 & 4 & 2.8 \\
\quad dataset-org/dream & dialogue question & 5{,}000 & 3 & 3 & 0.9 \\
\quad allenai/cosmos\_qa & narrative question & 10{,}000 & 4 & 4 & 1.2 \\
\quad tasksource/QuALITY & long-document question & 2{,}446 & 4 & 4 & 1.0 \\
\quad google/boolq & yes/no about a passage & 6{,}999 & 2 & 2 & 1.0 \\
\multicolumn{2}{@{}l}{\textit{Inference and rules}} & \textit{279{,}898} & & & \textit{19.2} \\
\quad nyu-mll/glue (MNLI) & premise--hypothesis relation & 264{,}981 & 3 & 3 & 15.9 \\
\quad tasksource/contract-nli & NDA supports a statement? & 918 & 3 & 3 & 0.1 \\
\quad tasksource/sharc & rule applies to a user? & 5{,}999 & 4 & 4 & 0.8 \\
\quad tasksource/Boardgame-QA & proved from rules? & 8{,}000 & 3 & 3 & 2.4 \\
\multicolumn{2}{@{}l}{\textit{Text and intent classification}} & \textit{89{,}426} & & & \textit{13.4} \\
\quad mteb/banking77$^\star$ & banking intent (77) & 9{,}426 & 77 & 77 & 2.1 \\
\quad clinc/clinc\_oos (plus)$^\star$ & assistant intent (151) & 5{,}000 & 149 & 151 & 1.3 \\
\quad CogComp/trec & question answer type & 5{,}000 & 50 & 50 & 0.6 \\
\quad fancyzhx/ag\_news & news topic & 10{,}000 & 4 & 4 & 0.7 \\
\quad fancyzhx/dbpedia\_14 & entity category & 20{,}000 & 14 & 14 & 2.1 \\
\quad community-datasets/yahoo\_answers\_topics & question topic & 15{,}000 & 10 & 10 & 1.1 \\
\quad BEE-spoke-data/consumer-finance-complaints & complaint product & 25{,}000 & 11 & 109 & 5.5 \\
\multicolumn{2}{@{}l}{\textit{Agents, tools and dialogue}} & \textit{67{,}649} & & & \textit{32.7} \\
\quad osunlp/Multimodal-Mind2Web$^\star$ & next web element & 3{,}000 & 6 & 64 & 8.4 \\
\quad bevaya/RICO-SCA & UI element for an instruction & 10{,}000 & 17 & 223 & 8.4 \\
\quad lockon/xlam-function-calling-60k & tool to call & 15{,}000 & 4 & 8 & 1.8 \\
\quad nvidia/When2Call & call, ask, refuse or answer & 9{,}649 & 4 & 4 & 2.4 \\
\quad samatv256/jev-decisions-v1 & agent's next option & 5{,}000 & 5 & 28 & 6.8 \\
\quad pfb30/multi\_woz\_v22 & dialogue intent & 25{,}000 & 7 & 255 & 4.8 \\
\multicolumn{2}{@{}l}{\textit{Response judging and safety}} & \textit{44{,}057} & & & \textit{10.1} \\
\quad nvidia/HelpSteer2 & response rating (0--4) & 34{,}057 & 5 & 5 & 9.2 \\
\quad nvidia/Aegis-AI-Content-Safety-Dataset-2.0 & is the request unsafe? & 10{,}000 & 2 & 2 & 0.9 \\
\multicolumn{2}{@{}l}{\textit{Synthetic procedural decisions}} & \textit{69{,}989} & & & \textit{12.1} \\
\quad tasksource/procedural-typed-decisions & 13 configurations$^{a}$ & 69{,}989 & 2--7 & 59 & 12.1 \\
\midrule
Total & & 891{,}107 & & & 108 \\
\bottomrule
\end{tabular}
\par\smallskip\raggedright\scriptsize $^{a}$ arithmetic, entity belief tracking, event state reconstruction, evidence sufficiency, multi view adjudication, needle retrieval, partial observation calibration, policy applicability, policy under uncertainty, record aggregation, state perturbation, table lookup, taxonomy routing.
\end{table*}

%% file: F_latency.tex
\section{Latency protocol}
\label{app:latency}
We time every method on one NVIDIA RTX A6000 GPU in bf16 with full inputs.
All systems are 1.7B models on the same Qwen3 base; the baseline model and its ensembles run as a standard causal language model, and only \method{} uses its own attention mask.
Every method uses the same attention kernel, memory-efficient SDPA.
We use three request shapes: short requests (MMLU-Pro), many options (CLINC150, $K=151$) and long contexts (QuALITY \citep{pang2022quality}).
\emph{Latency} is the time to answer one question at batch size 1, reported as the median and 95th percentile over 100 questions (50 on QuALITY).
\emph{Throughput} is the number of questions answered per second when questions are batched under a budget of padded tokens, taking the best budget between 16k and 128k tokens.
Calibration and augmentation are not timed, since at inference they cost one pass, like the baseline model.
On the short-input sets, batch-1 latency varied by up to about 30\% between reruns on the shared machine, because such short calls are bound by kernel-launch overhead; on QuALITY it varied by a few percent.
\Cref{tab:latency} reports the results.

%% file: G_hosted_api.tex
\section{Hosted API details}
\label{app:api}
We query the Jev API (version 1.13.0) and the OpenAI Decisions API (its default model) through their public endpoints, with the same questions and Latin-square orders as the open models.
A question is scored only if the API returned a usable answer under every one of its orders; otherwise it is dropped for that API, and each API's metrics are computed on the questions it answered.
The Jev API answered every order of every question except one CLINC150 question.
The OpenAI Decisions API failed on 1.3\% of the questions across the evaluation sets, most often on long option lists (19 of 200 on MASSIVE with 60 options and 45 of 1,659 on MMLU-Pro) and never on Mind2Web.

%% file: H_precision.tex
\section{Precision}
\label{app:precision}
\Method{}, separate encoding and the filtered ensemble are order-invariant in exact arithmetic.
In bf16, however, the order changes where each option's tokens sit in the input and thus the order in which the GPU accumulates sums, which can change the rounding; when two options are nearly tied, this can flip the answer.
Such flips are rare: \method{} 1.7B changes its answer on at most 0.06\% of order pairs (on Mind2Web steps; 0.01\% on MMLU-Pro and none on the other sets), the 4B model on none, and in fp32 the answers do not change with the order.
We therefore still treat these variants as order-invariant, since such flips are rare in bf16 and absent in fp32. 
We therefore evaluate training seeds 1 and 2 of \method{} and the filtered ensemble in the original order only.
Separate encoding is evaluated in the original order only for all three seeds, so its zero accuracy drop and flip rate hold by construction rather than being measured.

%% file: I_boxplots.tex
\section{Worst-case accuracy against token cost}
\label{app:boxplots}

\begin{figure*}[h]
  \centering
  \includegraphics[width=\textwidth]{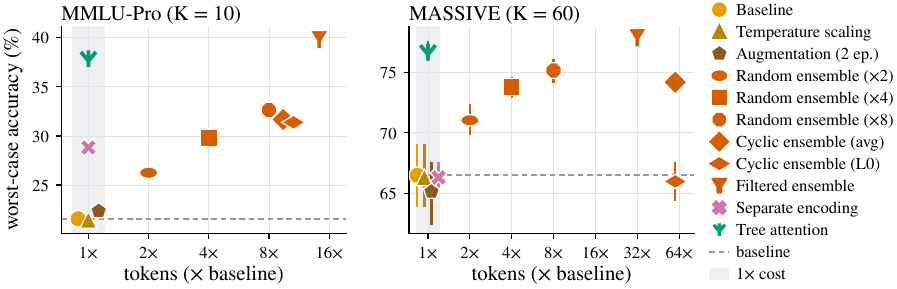}
  \caption{Worst-case accuracy vs.\ the token cost relative to the baseline model, the accuracy counterpart of \cref{fig:4}.
  Marker = mean over 3 training seeds; bar = minimum to maximum over the seeds; dashed line = baseline model. At $1\times$ only the method with the highest worst-case accuracy of each family is shown. Overlapping points at the same $x$ are spread slightly, centred on their $x$.}
  \label{fig:methods-acc}
\end{figure*}

%% file: J_separate_encoding.tex
\section{Observed failure modes in separate option encoding}
\label{app:sep}
\input{appJ_sep_patterns}
\input{appJ_sep_examples}

Separate encoding is order-invariant but loses accuracy on most sets (\cref{sec:res-sensitivity}).
Because each option is encoded without the question, the read-out can only compare a summary of the question with a summary of each option.
We check which questions it loses on, using the 1.7B models in the original option order, averaged over 3 training seeds, against the baseline and \method{}, whose options attend to the question.

\para{Question patterns.}
On MMLU, MMLU-Pro and ARC we mark three patterns by simple text rules (a question can match more than one).
A question is \emph{negated} if it contains NOT, EXCEPT or LEAST in capitals, or if its last sentence contains ``except'' or which/what/who/whom followed by a negative word (not, least, incorrect, false, cannot, isn't, \ldots).
It has a \emph{computed answer} if every option is a number (with an optional short unit) and the question contains a digit.
It has a \emph{specific ask} if it is not negated and its last sentence contains a superlative or qualifier (most, best, largest, main, primary, first, \ldots).

\para{Negated questions show the largest gap} (\cref{tab:sep_patterns}, top).
Separate encoding answers \appJNegSep\% of the \appJNegN{} negated questions correctly, against \appJNegBase\% for the baseline and \appJNegMeth\% for \method{}: a gap of \appJNegGap{} points, against \appJAllGap{} on all questions.
Computed answers and specific asks show gaps slightly below the overall one.
A negation reverses which option fits, and an option encoded without the question is likely matched mainly to the question's topic, so separate encoding tends to pick a typical member of the category (X-rays for ``does NOT belong in the family of electromagnetic waves'', \cref{tab:sep_examples}).

\para{The negation gap is not chance.}
Random question sets of the same size and the same mix of MMLU, MMLU-Pro and ARC questions (\appJDraws{} draws) have a mean gap of \appJNullMean{} points (95\% range \appJNullLo--\appJNullHi), against \appJNegGap{} for the negated questions (\appJNullP).
Per set (\cref{tab:sep_patterns}, middle), the gap on negated questions is larger than on the other questions of the same set in all three sets; it is significant on MMLU and on ARC (in-domain), but not on MMLU-Pro, which has only \appJProNegN{} negated questions.

\para{On MASSIVE, separate encoding often finds the domain but not the intent} (\cref{tab:sep_patterns}, bottom).
It reaches \appJMassSep\% accuracy, against \appJMassBase\% for the baseline and \appJMassMeth\% for \method{}.
Of its \appJMassErr{} errors (over 3 seeds), \appJMassSame\% pick a label from the gold label's domain, although only \appJMassBaseRate\% of all wrong labels lie there; the share is lower for the baseline and \method{}.
Separate encoding thus confuses near-identical labels that differ in the one word of the request that decides them (``alarm query'' vs.\ ``alarm set'').

\para{The decision head and the implementation do not change the negation gap.}
The ablation with Jeva's decision head and the control with Jeva's own implementation answer \appJNegAbl\% and \appJNegCtrl\% of the negated questions correctly, close to separate encoding's \appJNegSep\% (baseline \appJNegBase\%), which points to encoding the options separately as the cause.

\para{Examples.}
\Cref{tab:sep_examples} lists two questions per pattern that the baseline answers correctly in every option order and \method{} answers correctly, on all 3 seeds, while separate encoding misses them on all 3 seeds.
They illustrate the patterns; they are not a random sample.

%% file: appJ_sep_patterns.tex
\newcommand{\appJNegN}{205}
\newcommand{\appJNegSep}{44.4}
\newcommand{\appJNegBase}{68.0}
\newcommand{\appJNegMeth}{68.1}
\newcommand{\appJNegGap}{23.6}
\newcommand{\appJOtherGap}{10.7}
\newcommand{\appJAllGap}{11.3}
\newcommand{\appJNullMean}{11.6}
\newcommand{\appJNullLo}{6.0}
\newcommand{\appJNullHi}{17.4}
\newcommand{\appJNullP}{$p<0.001$, 0 of 20{,}000 draws}
\newcommand{\appJDraws}{20,000}
\newcommand{\appJMassSep}{66.3}
\newcommand{\appJMassBase}{78.5}
\newcommand{\appJMassMeth}{76.7}
\newcommand{\appJMassErr}{202}
\newcommand{\appJMassSame}{50.5}
\newcommand{\appJMassBaseRate}{5.7}
\newcommand{\appJNegAbl}{43.9}
\newcommand{\appJNegCtrl}{44.2}
\newcommand{\appJAllAbl}{46.3}
\newcommand{\appJAllCtrl}{46.4}
\newcommand{\appJGapAbl}{24.1}
\newcommand{\appJGapCtrl}{23.7}
\newcommand{\appJProNegN}{38}
\newcommand{\appJAllGapAbl}{13.5}
\newcommand{\appJAllGapCtrl}{13.4}
\begin{table}[t]
\centering
\caption{Separate encoding by question pattern. Accuracy (\%) in the original order, mean of 3 training seeds; MMLU, MMLU-Pro and ARC pooled. Top: per pattern; gap = baseline $-$ separate encoding (points). Middle: negated questions per set; the gap on negated vs.\ other questions of the set and the one-sided $p$ of a permutation test (20,000 random same-size question sets; pooled row stratified by set; $<$0.001: none of the draws reached the observed gap). Bottom: MASSIVE (60 intents); errors summed over 3 seeds; same domain = the wrong label shares the gold label's domain (its first word).}
\label{tab:sep_patterns}
\footnotesize
\setlength{\tabcolsep}{3.5pt}
\begin{tabular}{lrrrrr}
\toprule
Pattern & $n$ & Separate & Baseline & \Method{} & Gap \\
\midrule
Negated question & 205 & 44.4 & 68.0 & 68.1 & 23.6 \\
Computed answer & 675 & 34.1 & 42.6 & 42.1 & 8.5 \\
Specific ask & 1035 & 56.7 & 66.5 & 67.6 & 9.8 \\
\midrule
All questions & 4733 & 48.4 & 59.8 & 59.4 & 11.3 \\
\bottomrule
\end{tabular}

\vspace{0.6em}
\begin{tabular}{lrrrrrr}
\toprule
Negated, per set & $k/n$ & Separate & Baseline & Gap & Gap other & $p$ \\
\midrule
MMLU & 115/1909 & 49.0 & 68.4 & 19.4 & 11.3 & 0.026 \\
MMLU-Pro & 38/1659 & 21.1 & 36.0 & 14.9 & 9.9 & 0.256 \\
ARC (in-domain) & 52/1165 & 51.3 & 90.4 & 39.1 & 11.2 & $<$0.001 \\
\midrule
Pooled & 205/4733 & 44.4 & 68.0 & 23.6 & 10.7 & $<$0.001 \\
\bottomrule
\end{tabular}

\vspace{0.6em}
\begin{tabular}{lrrr}
\toprule
MASSIVE & Separate & Baseline & \Method{} \\
\midrule
Accuracy & 66.3 & 78.5 & 76.7 \\
Errors & 202 & 129 & 140 \\
Same domain (\%) & 50.5 & 35.7 & 45.7 \\
\bottomrule
\end{tabular}
\end{table}

%% file: appJ_sep_examples.tex
\begin{table}[t]
\centering
\caption{Questions that the baseline (in every option order) and \method{} answer correctly on all 3 seeds and separate encoding misses on all 3 seeds, two per pattern. Questions shortened; MASSIVE shows the request. Separate encoding's picks over its 3 seeds, the most frequent first.}
\label{tab:sep_examples}
\footnotesize
\setlength{\tabcolsep}{3pt}
\begin{tabular}{@{}p{0.52\linewidth}p{0.18\linewidth}p{0.24\linewidth}@{}}
\toprule
Question & Gold & Separate encoding \\
\midrule
\multicolumn{3}{@{}l}{\textit{Negation}} \\
Which of these does NOT belong in the family of electromagnetic waves? {\scriptsize (MMLU)} & Sound & X-rays \\
Which of these is not part of an atom? {\scriptsize (ARC)} & isotope & proton / electron \\
\midrule
\multicolumn{3}{@{}l}{\textit{Computed answer}} \\
Write 27 over 36 in simplest form. {\scriptsize (MMLU-Pro)} & 3 over 4 & 27 over 36 \\
Samantha has 342 baseball cards. Perry has 184 baseball cards. How many more baseball cards\dots{} {\scriptsize (MMLU-Pro)} & 158 & 178 / 168 \\
\midrule
\multicolumn{3}{@{}l}{\textit{Specific ask}} \\
What is the most abundant fossil fuel? {\scriptsize (MMLU)} & Coal & Petroleum \\
In America what is the main ingredient in pepperoni? {\scriptsize (MMLU)} & meat & pepper \\
\midrule
\multicolumn{3}{@{}l}{\textit{MASSIVE intent classification}} \\
``check if alarm is set for six am'' {\scriptsize (MASSIVE)} & alarm query & alarm set \\
``how is my order'' {\scriptsize (MASSIVE)} & takeaway query & takeaway order \\
\bottomrule
\end{tabular}
\end{table}